\documentclass{article}

\usepackage{microtype}
\usepackage{graphicx}
\usepackage{booktabs} 
\usepackage{longtable} 
\usepackage{verbatim}
\usepackage{subcaption}

\usepackage{hyperref}

\usepackage[accepted]{icml2024}

\usepackage{amsmath}
\usepackage{amssymb}
\usepackage{mathtools}
\usepackage{amsthm}

\usepackage[capitalize,noabbrev]{cleveref}

\theoremstyle{plain}

\theoremstyle{definition}

\theoremstyle{remark}

\usepackage{tabularray}
\usepackage{multirow}
\usepackage{array, booktabs}
\newcolumntype{C}[1]{>{\centering\let\newline\\\arraybackslash\hspace{0pt}}m{#1}}
\newcolumntype{M}[1]{>{\centering\arraybackslash}m{#1}}
\usepackage{anyfontsize}

\usepackage{color, colortbl}
\definecolor{Gray}{gray}{0.9}
\definecolor{DarkGray}{rgb}{0.75, 0.75, 0.75}
\definecolor{LightCyan}{gray}{0.95}
\newcolumntype{g}{>{\columncolor{Gray}}c}
\usepackage[textsize=tiny]{todonotes}

\icmltitlerunning{SLEB: Streamlining LLMs through Redundancy Verification and Elimination of Transformer Blocks}

\begin{document}

\twocolumn[
\icmltitle{SLEB: Streamlining LLMs through Redundancy Verification and\\
Elimination of Transformer Blocks}



\icmlsetsymbol{equal}{*}

\begin{icmlauthorlist}
\icmlauthor{Jiwon Song}{equal,snu}
\icmlauthor{Kyungseok Oh}{equal,snu}
\icmlauthor{Taesu Kim}{sqb}
\icmlauthor{Hyungjun Kim}{sqb}
\icmlauthor{Yulhwa Kim}{skku}
\icmlauthor{Jae-Joon Kim}{snu}
\end{icmlauthorlist}

\icmlaffiliation{snu}{Seoul National University}
\icmlaffiliation{sqb}{SqueezeBits Inc.}
\icmlaffiliation{skku}{Sungkyunkwan University}

\icmlcorrespondingauthor{Yulhwa Kim}{yulhwakim@skku.edu}
\icmlcorrespondingauthor{Jae-Joon Kim}{kimjaejoon@snu.ac.kr}

\icmlkeywords{Machine Learning, ICML}

\vskip 0.3in
]



\printAffiliationsAndNotice{\icmlEqualContribution} 

\begin{abstract}
Large language models (LLMs) have proven to be highly effective across various natural language processing tasks. However, their large number of parameters poses significant challenges for practical deployment.
Pruning, a technique aimed at reducing the size and complexity of LLMs, offers a potential solution by removing redundant components from the network. Despite the promise of pruning, existing methods often struggle to achieve substantial end-to-end LLM inference speedup.
In this paper, we introduce SLEB, a novel approach designed to streamline LLMs by eliminating redundant transformer blocks.
We choose the transformer block as the fundamental unit for pruning, because LLMs exhibit block-level redundancy with high similarity between the outputs of neighboring blocks. This choice allows us to effectively enhance the processing speed of LLMs.
Our experimental results demonstrate that SLEB outperforms previous LLM pruning methods in accelerating LLM inference while also maintaining superior perplexity and accuracy, making SLEB as a promising technique for enhancing the efficiency of LLMs.
The code is available at: \url{https://github.com/jiwonsong-dev/SLEB}.
\end{abstract}

\section{Introduction}
\label{sec:intro}

Large language models (LLMs), such as GPT-3, OPT, and LLaMA demonstrate exceptional proficiency in a variety of natural language processing (NLP) tasks and have become key components in applications like chatbots and question-answering systems~\cite{gpt3, opt, palm, llama_v1, llama_v2}.
However, their substantial number of parameters creates significant challenges in deploying these models for real-world services, especially due to the increased memory consumption and computational demands.
This limitation restricts their widespread use. 
Consequently, it is critical to develop techniques that improve the compactness and processing efficiency of LLMs while preserving their linguistic prowess.

Network pruning is a technique aimed at reducing the size and complexity of neural networks by eliminating redundant weight parameters~\cite{obd, obs, songhan}. Its application in LLMs has been somewhat limited, primarily due to challenges that arise in managing sparse matrices~\cite{obc, wanda, sparsegpt, DSnoT}.
This complexity becomes particularly evident when using modern GPU hardware, as these systems are typically optimized for operations involving dense matrices~\cite{sparsert, sparsegpu}.

\begin{figure}[t!]
    \centering
    \includegraphics[width=0.72\columnwidth]{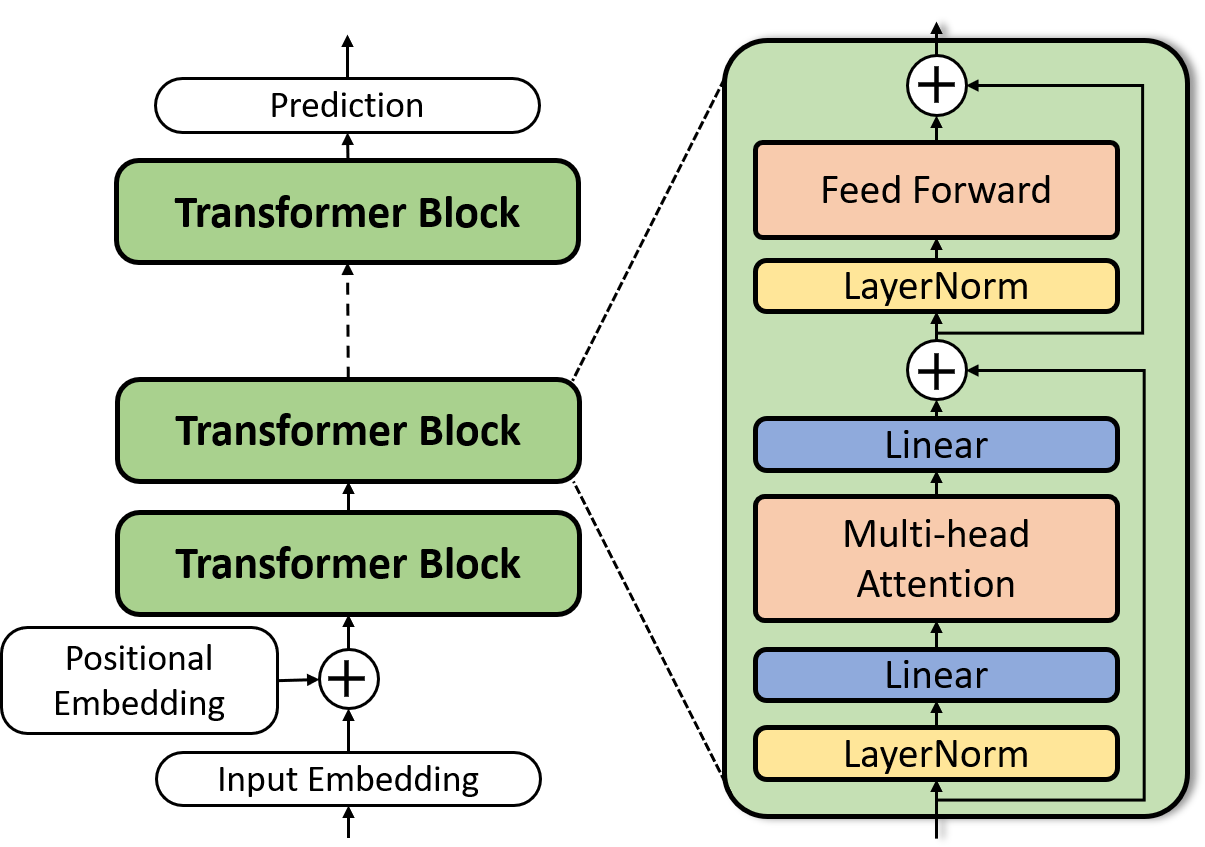}
    \vspace{-3mm}
    \caption{Typical LLM architecture}
    \label{fig:llm}
    \vspace{-6mm}
\end{figure}

\begin{figure*}[t]
    \centering
    \includegraphics[width=0.8\textwidth]{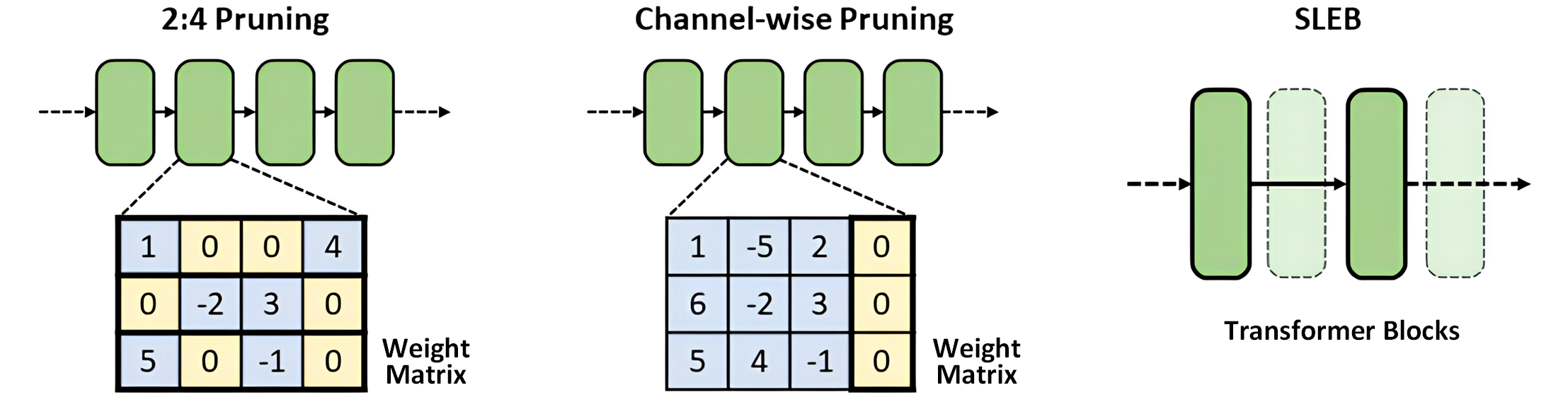}
    \vspace{-4mm}
    \caption{Overview of previous pruning methods, 2:4 pruning and channel-wise (a.k.a. row/column-wise) pruning, and proposed SLEB}
    \label{fig:overview}
    \vspace{-4mm}
\end{figure*}

In the realm of LLMs, a significant similarity in output is observed among successive transformer blocks~\cite{jump_to_con, dejavu}. This similarity arises because each transformer block incrementally contributes to the residual path spanning the entire LLM.
Figure~\ref{fig:llm} depicts the typical architecture of conventional LLMs, characterized by a continuous stack of transformer blocks. A key aspect of LLM computation is the residual path that traverses the entire network, a feature introduced to stabilize the backpropagation process during training.
Consequently, each transformer block contributes its computational outputs, derived from both the attention mechanisms and feed-forward layers, to this residual path.
This design feature results in a considerable degree of output similarity between consecutive transformer blocks, leading to redundancy within the LLMs.


In this paper, we propose SLEB, a novel approach designed to streamline LLMs by identifying and eliminating redundant transformer blocks.
Figure~\ref{fig:overview} compares the proposed approach with previous pruning methods.
SLEB is tailored to refine LLMs through the strategic removal of redundant transformer blocks, thereby effectively aligning speedup with the pruning ratio. 
The fundamental principle behind SLEB is that the careful elimination of redundant transformer blocks can be achieved without affecting the text generation capabilities of LLMs.
By targeting these specific redundant elements within the LLM's architecture, SLEB seeks to provide a more efficient pruning approach. It aims to overcome the challenges typically associated with traditional network pruning, particularly in enhancing the acceleration of LLMs.

\section{Motivation}
\label{sec:motivation}

\subsection{Pruning}
\label{sec:pruning}

Standard LLMs are designed as shown in Figure~\ref{fig:llm}, with a series of identically structured transformer blocks arranged sequentially. Each transformer block in these models consists of attention and feed-forward layers.
The operations in both the linear and feed-forward layers primarily involve matrix multiplication between the weight parameters and input activations. Given this structure, there are two main strategies for developing compact and fast LLMs: enhancing the efficiency of each individual transformer block or reducing the overall number of blocks.
To improve both the compactness and processing speed of individual blocks, pruning is frequently utilized~\cite{obd, obs, songhan}. This technique involves the strategic removal of superfluous weight parameters, thereby reducing the computational load associated with matrix multiplication in each linear and feed-forward layer.
However, the application of pruning techniques to LLMs encounters a notable challenge.

\textbf{Challenge 1) Limitation in Achieving LLM Inference Speedup:} Pruning can be divided into two main types: unstructured and structured~\cite{obc, wanda, sparsegpt, DSnoT}. Unstructured pruning targets the removal of individual weights, leading to sparse weight matrices within the model. This sparsity can create complex data access patterns, complicating the management of sparse data and potentially hindering the model's acceleration. Particularly with NVIDIA GPUs, achieving a speedup through unstructured pruning typically requires reaching a high level of sparsity, often more than 90\%~\cite{sparsert, sparsegpu}. However, for LLMs, achieving pruning ratios above 50\% is often difficult without substantially compromising their linguistic prowess.
Structured pruning, on the other hand, involves the elimination of predefined units of weights to create more hardware-friendly patterns, such as 2:4 pruning~\cite{wanda, sparsegpt, DSnoT} or channel-wise (a.k.a. row/column-wise) pruning~\cite{slicegpt, llm_pruner}.
The goal of structured pruning is to form dense matrices that are more efficiently processed by hardware, thereby enhancing efficiency.
Nonetheless, the anticipated speedup from structured pruning, ideally proportional to the pruning ratio, frequently fails to be realized in practical settings.

\begin{figure}[t]
    \centering
    \includegraphics[width=0.80\columnwidth]{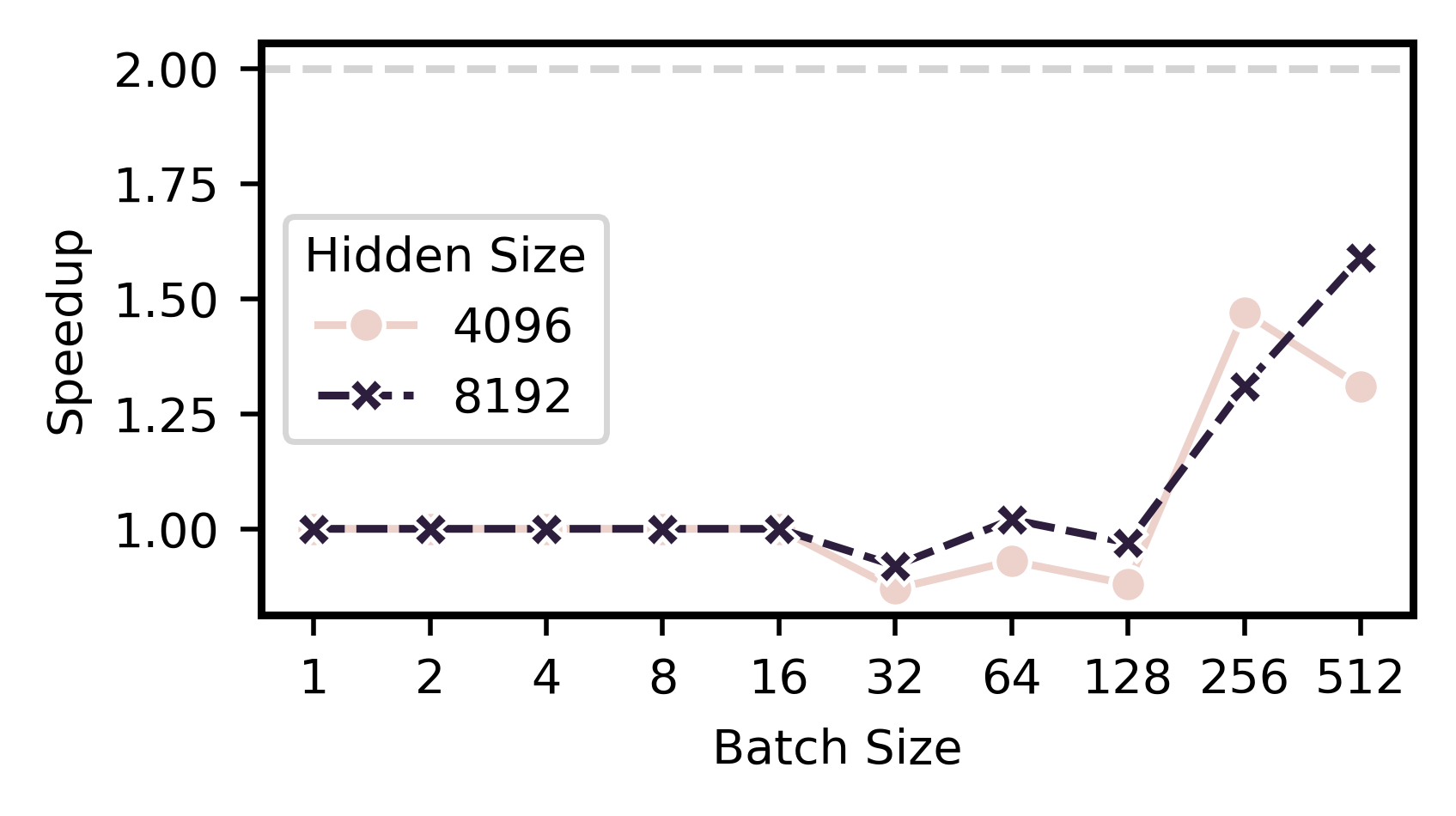}
    \vspace{-6mm}
    \caption{The speedup achieved through 2:4 pruning on matrix multiplication between $b \times h$ matrix and $h \times h$ ($b$: batch size, $h$: hidden size).
    The test is conducted on an NVIDIA RTX A6000 GPU.
    The dashed grey line represents the peak speedup attainable with 2:4 pruning.
    The speed of the 2:4 pruning case with batch sizes between 1 and 16 is measured using a dense matrix multiplication kernel, because NVIDIA CUTLASS effectively supports 2:4 sparse matrix multiplication with batch sizes larger than 16.
    }
    \label{fig:pruning_matmul_speedup}
    \vspace{-4mm}
\end{figure}

Recent NVIDIA GPUs support the acceleration of 2:4 fine-grained structured sparsity using their Tensor Cores~\cite{24gpu}.
This form of sparsity involves having two zero values within each contiguous block of four values, resulting in a natural sparsity rate of 50\%. The peak performance of sparse tensor cores is twice that of dense tensor cores. 
Thus, previous research on pruning LLMs has extensively examined 2:4 pruning techniques~\cite{wanda, sparsegpt, DSnoT}.
However, it is challenging to achieve the desired speedup with 2:4 pruning.
The speedup achieved with 2:4 pruning heavily depends on the size of the input matrix. 
Figure~\ref{fig:pruning_matmul_speedup} provides a comparative analysis of matrix multiplication latency on an NVIDIA RTX A6000 GPU, using both NVIDIA cuBLAS and CUTLASS libraries. The matrix multiplication involves batched inputs with dimensions of $b \times h$, along with a weight matrix size of $h \times h$, where $b$ represents the batch size and $h$ represents the hidden size.
The measurement results indicate that within a realistic batch size range of 1 to 128, 2:4 pruning does not result in a speedup. However, as the batch size is increased beyond this range, 2:4 pruning begins to show a speedup, eventually reaching approximately 1.5$\times$ the speed of dense matrix multiplication. To achieve significant speedups, it is essential to have a sufficiently large matrix size, which can lead to high arithmetic intensity~\cite{24gpu}.
In summary, achieving faster LLM inference with conventional pruning methods has two main limitations. Firstly, it requires dedicated hardware, such as the 2:4 sparse matrix multiplication accelerator adapted in recent NVIDIA GPUs, to enhance inference speed. Secondly, even with dedicated hardware support, the extent of speedup achievable through 2:4 pruning is influenced by factors like batch size and model size.

\begin{figure}[t]
    \centering
    \includegraphics[width=\columnwidth]{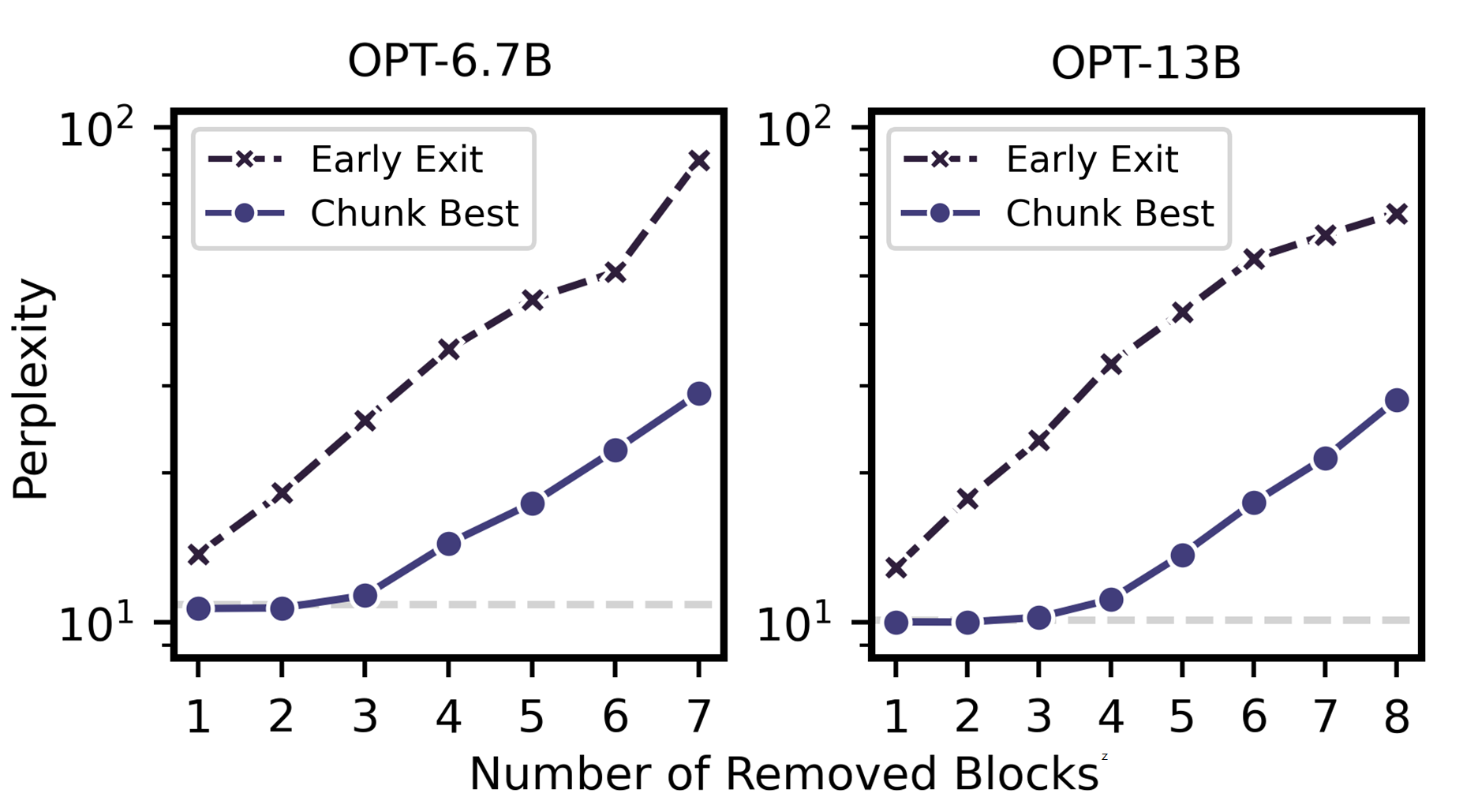}
    \vspace{-9mm}
    \caption{Perplexity comparison on WikiText-2 for OPT-6.7B (left) and OPT-13B (right) after removing consecutive transformer blocks. ``Early Exit" refers to removing the very last blocks from the target model, while ``Chunk Best" represents the best perplexity results achieved by testing all possible removable points of consecutive blocks.}
    \label{fig:chunk_ppl}
    \vspace{-6mm}
\end{figure}

As an alternative to traditional pruning methods, approaches like LLM-Pruner~\cite{llm_pruner} and SliceGPT~\cite{slicegpt} have been developed, focusing on the elimination of redundant neurons and the weight parameters connected to them.
These methods involve the elimination of the entire channel (row/column) from weight matrices, allowing pruned weights to remain in a dense format. This retention of dense matrices enables direct speedup in matrix multiplication.
However, it is challenging to maintain linguistic capabilities of LLMs with these channel-wise pruning approaches without the help of extensive fine-tuning~\cite{llm_pruner}.
Additionally, speedup in matrix multiplication does not always lead to corresponding improvement in the end-to-end inference speed of LLMs~\cite{slicegpt}.
This limitation arises because transformer blocks in LLMs encompass various operations beyond matrix multiplication that involve weight parameters. 
These operations include, but are not limited to, layer normalization and attention mechanisms (Figure~\ref{fig:llm}).

Recently, Deja Vu~\cite{dejavu} has introduced a new pruning approach termed contextual sparsity.
This method dynamically assesses whether to bypass certain segments of a layer operation, based on the context of the input. It has shown effectiveness in accelerating inference, particularly in single-batch scenarios.
However, its dynamic nature introduces a set of challenges akin to those encountered with early exit strategies. These challenges will be further examined and discussed in the following subsection.

\subsection{Early Exit}

\begin{figure}[t]
    \begin{minipage}{0.5\textwidth}
        \centering
        \includegraphics[width=\columnwidth]{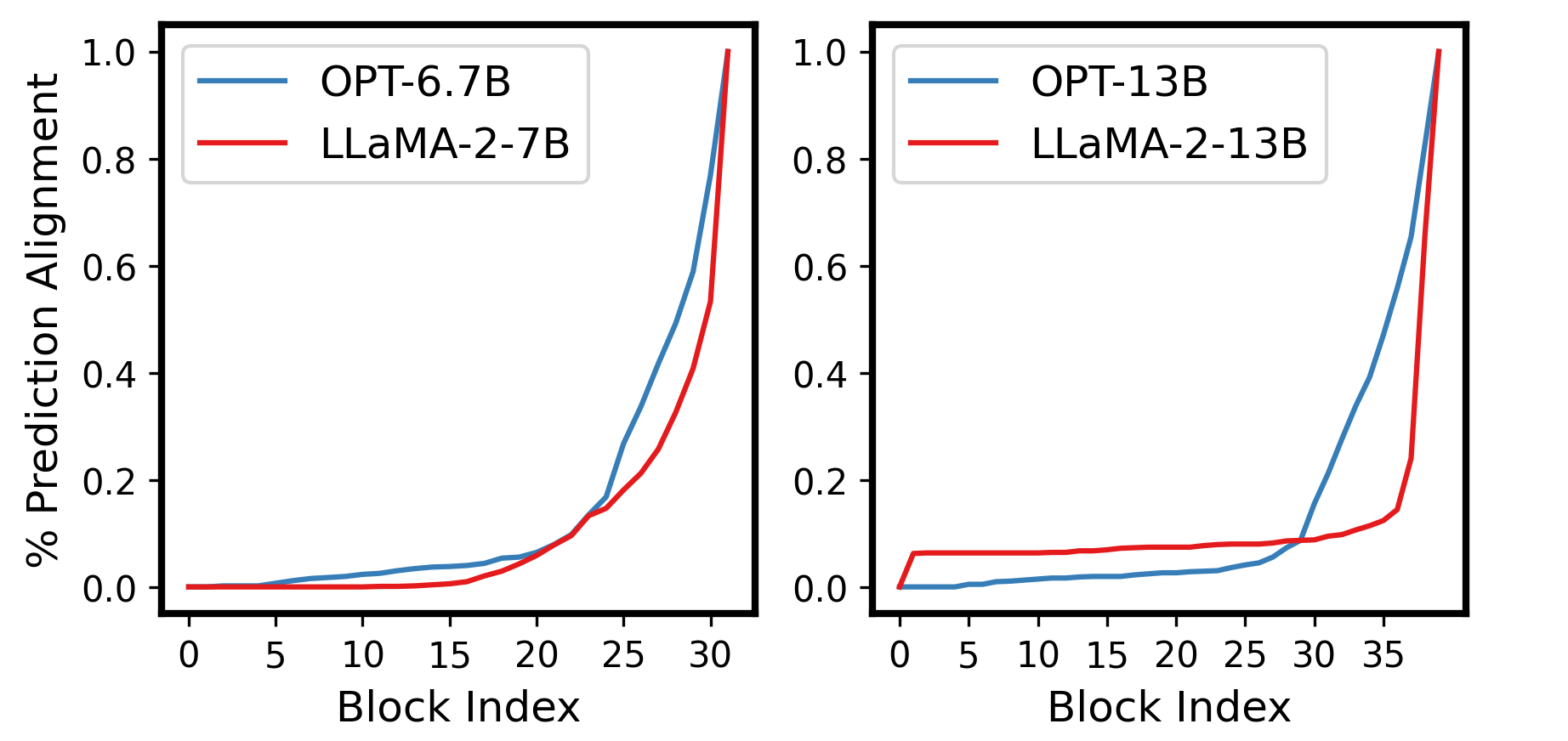}
        \vspace{-9mm}
        \caption{Percentage of token prediction alignment with the final predictions of LLMs for each transformer block.}
        \label{fig:early_exit}
        \vspace{-6mm}
    \end{minipage}
\end{figure}

Reducing the total number of transformer blocks in the LLM can directly lead to increased processing speed, proportional to the reduction in the number of blocks.
A notable technique in this context is the early exit~\cite{calm, jump_to_con, lite, ee_llm}. This method dynamically assesses whether to process subsequent transformer blocks. 
When the model reaches a certain level of confidence in its early outputs, it can halt further processing, effectively reducing computational cost.
Another recent approach~\cite{skipdecode} proposes an alternative method for bypassing transformer blocks. This strategy suggests that removing blocks, particularly those at the beginning of LLMs, might be more feasible.
However, all of these strategies, which involve skipping transformer blocks, typically require either dynamic decision-making or extensive training to be effective.
To assess the impact of such removals without dynamic decision-making and training, we analyze the perplexity scores of OPT-6.7B/13B on the WikiText-2 dataset~\cite{wikitext2} after removing various consecutive blocks, as illustrated in Figure~~\ref{fig:chunk_ppl}.
The analysis results show that the ``Early Exit" approach leads to a noticeable decline in the linguistic prowess of the LLMs, as evidenced by a significant increase in their perplexity scores.
Moreover, ``Chunk Best", which represents the best perplexity results achieved by testing all possible removable points of consecutive blocks, also results in a significant increase in perplexity as the number of removed blocks increases.
Therefore, the concept of removing consecutive blocks from LLMs is not effective without dynamic decision-making and training. In addition, applying early exit to LLMs, particularly in real-world implementation, faces the following challenges.

\textbf{Challenge 2) Limitation in Acceleration in Multi-batch Settings:} In practical LLM applications, multi-batch scenarios are common and pose significant challenges. Processing individual tokens can lead to diverse dynamic skipping decisions. When these tokens are batched for parallel processing, each token may require different processing paths. This variation complicates the implementation and diminishes the efficiency benefits in multi-batch contexts.

\textbf{Challenge 3) Inability to Reduce Memory Requirements:} Dynamic methods such as early exit face limitations in reducing memory consumption, as they require maintaining the entire set of model parameters. This limitation is particularly impactful for models with a large number of parameters, where memory efficiency is crucial.

\textbf{Challenge 4) Resource-Intensive Training:}
As shown in Figure~\ref{fig:early_exit}, transformer blocks that produce the same prediction results as the final token prediction of the LLM are typically positioned near the end of the model.
Even with the use of an ideal skipping predictor for accurately determining an early exit point, it is estimated that around 90\% of the transformer blocks would still require processing. 
Hence, implementing early exit strategies demands extensive training. 
This process is not limited to training the skipping predictor but also involves adapting the LLMs to effectively utilize early exit mechanisms.
Such a training process can be resource-heavy and complicated.
Owing to these challenges, the implementation of early exit strategies in state-of-the-art LLMs, particularly those with more than 13 billion parameters, has not been extensively pursued. This situation emphasizes the significant difficulties in adapting these methods for use in extremely large-scale models like LLaMA-2-70B.

In the upcoming section, we introduce SLEB, a novel approach designed to streamline LLMs by identifying and eliminating redundant transformer blocks.
SLEB is developed to address challenges faced by previous works in the following ways:

\textbf{Solution to Challenge 1) LLM Inference Speedup:} SLEB adopts transformer block as the primary unit for elimination, aiming to achieve a speedup that is directly proportional to the number of blocks removed.

\textbf{Solution to Challenge 2) Acceleration in Multi-batch Settings:} Unlike the early exit strategy, SLEB employs a static approach to eliminate transformer blocks following an in-depth verification of redundancy, aligning with traditional pruning techniques. This allows SLEB to enhance processing speed in multi-batch scenarios effectively.

\textbf{Solution to Challenge 3) Reduction in Memory Requirements:} By completely removing redundant blocks, SLEB can significantly decrease memory usage, contributing to more efficient memory management in LLMs.

\textbf{Solution to Challenge 4) Training-free Compression:} Relying on a training-free redundancy analysis, SLEB eliminates the need for intensive training processes. This makes it an especially suitable solution for large-scale models, where the demands and constraints of extensive training are often prohibitive.

\section{Proposed SLEB}
\label{sec:sleb}

\subsection{Output Similarity across Transformer Blocks}
\label{sec:similarity}

\begin{figure}[t]
    \centering
    \includegraphics[width=\columnwidth]{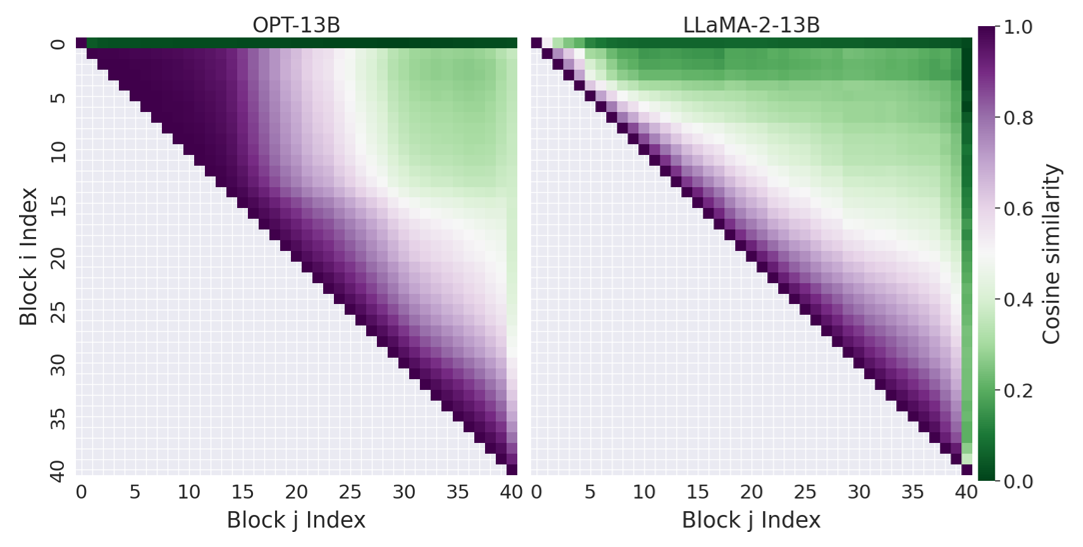}
    \vspace{-10mm}
    \caption{Cosine similarity between the outputs of two transformer blocks, Block $i$ and Block $j$.}
    \label{fig:cosin_sim}
    \vspace{-3mm}
\end{figure}


During the inference process of LLMs, there is a notable similarity in the outputs of consecutive transformer blocks. This similarity arises because each block incrementally adds to the residual path that runs throughout the entire LLM.
The computation results of each transformer block, considering the residual path, can be simplified using the following equation: 
\begin{equation}
    x_{i+1} = T_{i+1}(x_i) + x_i
    \label{eq:skip}
\end{equation}
Here, $x_i$ denotes the output of the $i$-th transformer block, and $T_i$ denotes the operation results of the $i$-th transformer block.
To comprehensively analyze and quantify the output similarity within the transformer blocks, we analyze cosine similarity between outputs of these blocks.
This analysis is based on the distance measurement method outlined in~\cite{jump_to_con}. It involves feeding a single random token into each LLM and measuring the cosine similarity between the output of the $i$-th transformer block and that of the $j$-th transformer block is defined as follows:
\begin{equation}
    similarity(x_i, x_j) = \frac{x_i \cdot x_j}{||x_i|| ||x_j||}
    \label{eq:dist}
\end{equation}
The measurement results for OPT-6.7B and LLaMA-2-7B models are displayed in Figure~\ref{fig:cosin_sim}.

These results reveal that, although the similarity between transformer blocks that are further apart varies across the LLM architecture, there is a consistently high degree of similarity between neighboring transformer blocks.
This insight underscores the potential redundancy within these models and highlights areas where efficiency could be enhanced through strategic block elimination. 

Indeed, this observation emphasizes a fundamental misalignment between existing early exit methods and the nature of LLMs. The primary limitation of early exit methods is their tendency to skip a continuous sequence of transformer blocks. This approach does not take into account the high degree of output similarity between adjacent transformer blocks, as indicated by our cosine similarity measurements. Consequently, such strategies risk bypassing blocks that are essential for preserving the linguistic integrity of the LLMs.

\subsection{Redundancy Verification of Transformer Blocks}
\label{sec:redundancy}


\begin{figure}[t]
    \centering
    \vspace{-2mm}
    \includegraphics[width=0.85\columnwidth]{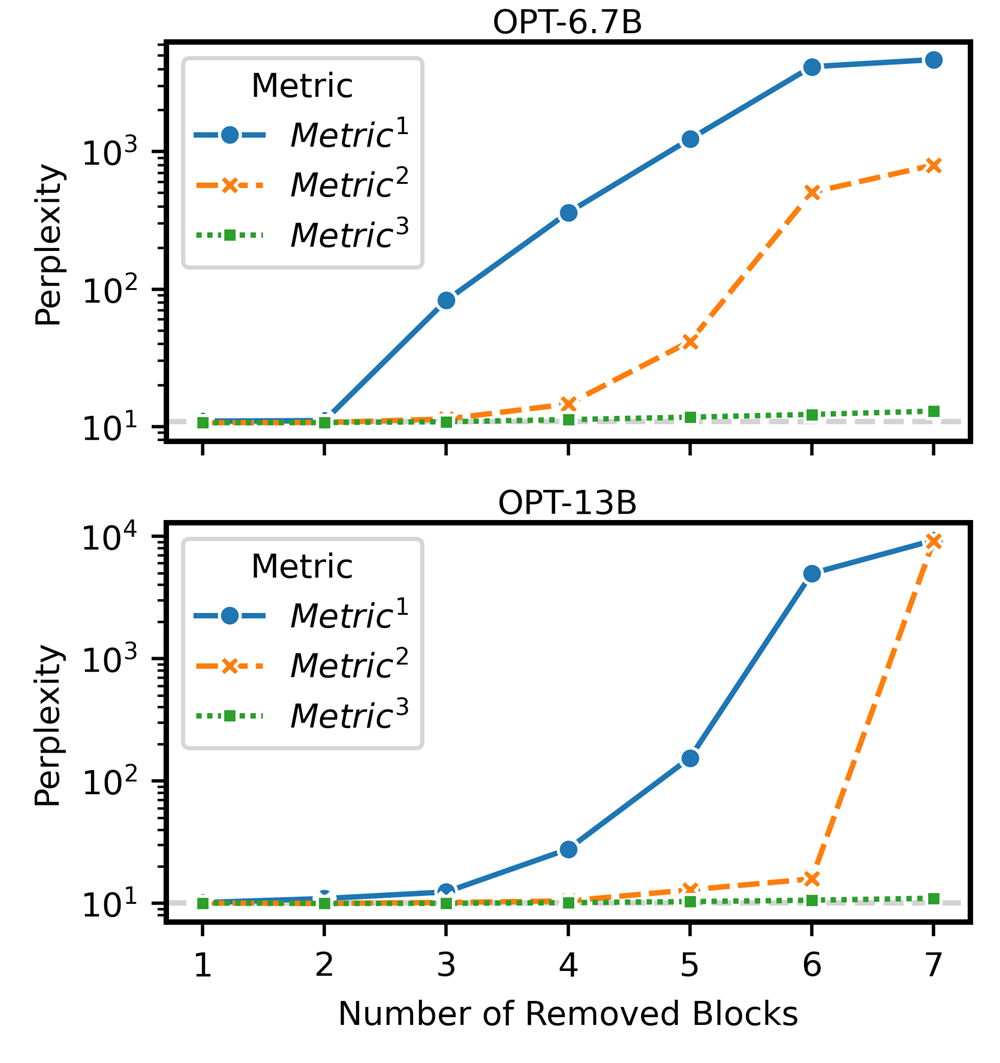}
    \vspace{-6mm}
    \caption{Perplexity comparison on WikiText-2 for OPT-6.7B (top) and OPT-13B (bottom) when transformer blocks are removed using different metrics.}
    \label{fig:metric_ppl}
    \vspace{-3mm}
\end{figure}

To streamline LLMs without affecting their linguistic prowess, it is essential to identify redundant transformer blocks first.
We design metrics to measure the significance of a block to eliminate blocks with lower metric values.
One potential metric is to assess the distance between the input and output of each block.
A small change after passing through the block might indicate a minor impact on the overall LLM inference.
In this context, we adopt Eq.~\ref{eq:dist} as the metric to measure the significance of transformer blocks.
Given the input and output of the $j$-th transformer block as $A_{j}$ and $B_{j}$, this metric can be defined as follows:
\begin{equation}
    Metric_{j}^{1} = 1 - similarity(A_j, B_j)
    \label{eq:metric1}
\end{equation}
$Metric^{1}$ is tested on removing transformer blocks from the OPT-6.7B and OPT-13B model, and the results are reported in Figure~\ref{fig:metric_ppl}.
However, using this metric led to a significant increase in perplexity. 
The reason for this increase is that even though a transformer block may have a minor impact on the residual path at its stage, as the LLM inference progresses, these minor changes in that block can be amplified, especially if the block lies in the early stage of the LLM, leading to a more substantial impact on the overall results.

To evaluate the significance of transformer blocks within the overall LLM inference process, we design a new metric to assess the impact of each block on the token prediction results of the target model.
We denote the target LLM model as $M$ and the LLM model after removing $j$-th transformer block as $M_j$. Using a tokenized sequence for the calibration as $w_{1}, w_{2},...,w_{K}$, the second metric for measuring the significance of $j$-th transformer block can be defined as follows:
\begin{equation}
\vspace{-2mm}
    Metric_{j}^{2} = -\frac{1}{K}\sum_{k=0}^{K}{log p_{M_{j}}(w_k|w_{<k})}
    \label{eq:metric2}
\end{equation}
$Metric^{2}$ shows significant improvements in perplexity on the block-removed LLMs compared to those removed using $Metric^{1}$ (Figure~\ref{fig:metric_ppl}).
However, as the number of removed blocks increases, $Metric^{2}$ also exhibits significant degradation in token prediction results, particularly for OPT-6.7B models.
This behavior can be explained by the observation such that the significance of a block keeps changing as other blocks are removed. For example, when using $Metric^{2}$, it selects blocks with indices 3, 4, 5, 6, 7, 8, and 10 for removal from OPT-6.7B. While these blocks individually have a minor impact on LLM inference results, the removal of a continuous sequence of blocks can significantly impact the overall LLM inference result.

To accurately assess the redundancy of transformer blocks and determine the next block for removal, we utilize the LLM with previously removed redundant blocks in an iterative removal process.
In this revised metric, we first evaluate the token prediction results of the original LLM to identify the most redundant block for removal. In the next stage, we assess the significance of each block with a block-removed LLM to select the next block for removal. This iterative process guides the removal of transformer blocks based on the current state of the LLM.
Denoting the target LLM model with previously removed redundant blocks as $M'$, the metric for identifying the next block for removal is as follows:
\begin{equation}
\vspace{-2mm}
    Metric_{j}^{3}(M') = -\frac{1}{K}\sum_{k=0}^{K}{log p_{M_{j}^{'}}(w_k|w_{<k})}
    \label{eq:metric3}
\end{equation}
When utilizing $Metric^{3}$ to assess the significance of each block and remove a total 7 of blocks from OPT-6.7B, blocks with the following indices are removed at each block removal step: 6, 7, 3, 24, 18, 30, 11.
It avoids the removal of a continuous sequence of blocks and successfully considers the changing significance of each block.
Consequently, as shown in Figure~\ref{fig:metric_ppl}, the LLMs streamlined using this metric maintain their perplexity scores remarkably well regardless of the target models.

\subsection{Proposed SLEB Algorithm}
\label{sec:sleb_algorithm}

\begin{algorithm}[tb]
\caption{SLEB algorithm. We remove the transformer blocks until the number of removed blocks reaches the target number.}
   \label{algo:sleb}
\begin{algorithmic}
    \STATE $M \longleftarrow original$ $model$
    \STATE $C \longleftarrow calibration$ $dataset$
    \STATE $N \longleftarrow \#$ $blocks$ $of$ $M$
    \STATE $n \longleftarrow \#$ $blocks$ $to$ $remove$
    \FOR{$i=0$ {\bfseries to} $n-1$}
    \FOR{$j=0$ {\bfseries to} $N-i-1$}
    \STATE $S\longleftarrow Metric_{j}^{3}(M, C)$
    \IF{$S<min\_S$}
    \STATE $min\_S \longleftarrow S$
    \STATE $min\_S\_idx \longleftarrow j$
    \ENDIF
    \ENDFOR
    \STATE $M\longleftarrow remove\_block(M,min\_S\_idx)$
    
    \ENDFOR
\end{algorithmic}
\vspace{-1mm}
\end{algorithm}

The proposed SLEB verifies and eliminates redundant transformer blocks with proposed $Metric^3$. The overall SLEB algorithm is summarized in Algorithm~\ref{algo:sleb}.
SLEB can be seamlessly integrated into the forward pass of the LLM model, similar to recent methods for pruning LLMs. It leverages calibration data to estimate the redundancy of transformer blocks and uses $Metric^3$ to pinpoint and remove the most redundant block. To determine the next block for removal, the algorithm calculates the redundancy of transformer blocks within the LLM that has been streamlined so far, allowing it to iteratively identify and prune redundant blocks. This approach streamlines the model without the need for an additional training process.

\section{Experiments}
\label{sec:experiments}

\subsection{Experimental Setup}

We implement SLEB in PyTorch~\cite{pytorch}, using the HuggingFace Transformers library~\cite{huggingface-transformers}.
The experiments on redundancy verification and elimination of transformer blocks are executed on NVIDIA A100 GPUs equipped with 80GB of memory. SLEB requires 2 GPUs for pruning OPT-66B and LLaMA-70B, and 1 GPU for pruning smaller models.
Runtime of SLEB for pruning LLMs falls within the range of runtime observed with other pruning methods (See Appendix~\ref{appendix:runtime}).
For example, SLEB can fully compress LLaMA-2-70B, within approximately 1.5 hours.
SLEB is processed based on the inference-only approach without any fine-tuning, aligning with contemporary practices in post-training pruning of LLMs. 
We use 128 samples randomly selected from WikiText-2 training dataset as calibration data, following the approach used in previous works~\cite{slicegpt}.

Our evaluation encompasses models from the OPT and LLaMA-2 families.
We assess SLEB under two target sparsity levels: 10\% and 20\%.
In case that the product of the total number of transformer blocks in a model and the target sparsity is not an integer, we round up the product to determine the number of transformer blocks to remove.
The proposed SLEB is compared to previous works that are widely recognized for their potential to accelerate LLM inference.
These previous works include SparseGPT~\cite{sparsegpt}, Wanda~\cite{wanda}, and DSnoT~\cite{DSnoT}, which utilize 2:4 structured pruning methods, and LLM-Pruner and SliceGPT, which employ channel-wise pruning.

\subsection{Elimination of Transformer Blocks using SLEB}
\begin{figure}[t]
    \vspace{-2mm}
    \centering
    \includegraphics[width=0.9\columnwidth]{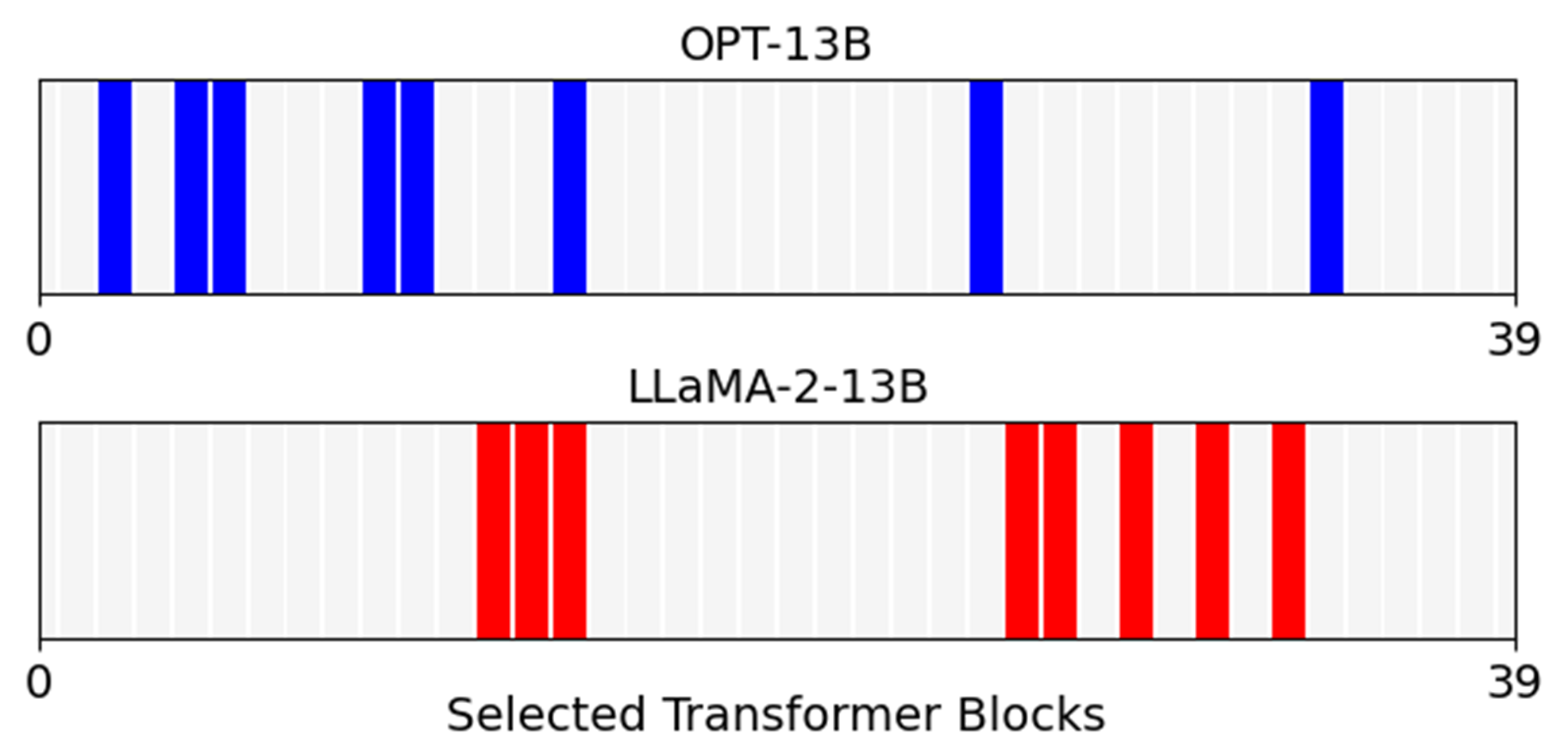}
    \vspace{-4mm}
    \caption{Locations of removed transformer blocks after applying SLEB with target sparsity of 20\%}
    \label{fig:selected}
    \vspace{-10mm}
\end{figure}

We assess which transformer blocks are chosen to be eliminated with SLEB in Figure~\ref{fig:selected}.
The locations of removed transformer blocks vary significantly across the target models.
In the case of OPT-13B, earlier transformer blocks are selected more frequently, while middle and later blocks are chosen for LLaMA-2-13B.
This demonstrates that the sensitivity to the removal of transformer blocks varies depending on the model.
Therefore, it is crucial to select a metric like the proposed $Metric^3$ that accurately reflects the characteristics of each model to identify the optimal combinations of transformer blocks for removal from various models.
The removed block indices for all evaluated LLMs are provided in Appendix~\ref{appendix:selected_blocks}.



\begin{table*}[t]
\caption{
Perplexity results on C4 dataset and throughput (tokens/s) results. We measure throughput of each method with LLaMA-2-70B on 2 NVIDIA A100 GPUs. (T.Block: Transformer Block)
}
\label{table:c4_ppl}
 \centering
  \footnotesize
\begin{tabular}{l | M{1.2cm} |  c | c | c | cccc | ccc}
\toprule
\rowcolor{Gray}
 ~ & \bf Pruning & ~ & ~ & \bf Throughput &\multicolumn{4}{c|}{\bf OPT} & \multicolumn{3}{c}{\bf LLaMA-2} \\
\rowcolor{Gray}
 \bf Method & \bf Unit & \bf Sparsity &  \bf  Tokens/s &  \bf Improve. &  \bf 6.7B & \bf  13B & \bf  30B & \bf  66B & \bf  7B & \bf  13B & \bf  70B \\
\midrule
\midrule
Dense		& - & 0\%  & 299 & 1.00$\times$ & 12.71	& 12.06	& 11.44	& 10.99		& 7.26		& 6.73	& 5.71 \\
\midrule
SparseGPT	& 2:4  & 50\%  & 293 & 0.98$\times$ & 16.42	& 14.85	& 13.17	& 12.25		& 13.54	& 11.39	& 8.16 \\
Wanda		& 2:4  & 50\%  & 293 & 0.98$\times$ & 19.03	& 16.18	& 16.18	& 8414.05	& 15.57	& 12.47	& 8.10	\\
DSnoT		& 2:4  & 50\%  & 293 & 0.98$\times$ & 18.41	& 16.51	& 14.71	& 8360.81	& 15.56	& 12.22	& 8.15	\\
\midrule
LLM-Pruner	& Channel & 20\% & 314 & 1.05$\times$ & -		& -		& -		& -			& 12.25	& 10.43	& -	\\
SliceGPT	& Channel & 20\% & 314 & 1.05$\times$ & 23.76	& 17.49	& 13.38	& 11.80		& 26.06	& 22.90	& 15.84	\\
SliceGPT	& Channel & 25\% & 331 & 1.11$\times$ & 27.35	& 19.43	& 14.46	& 12.29		& 32.74	& 29.86	& 20.03	\\
SliceGPT	& Channel & 30\% & 343 & 1.15$\times$ & 33.43	& 22.58	& 15.89	& 13.08    	& 41.69	& 38.43	& 25.79	\\
\midrule
\rowcolor{LightCyan}
SLEB		& T. Block & 10\% & 336 & 1.12$\times$ & 13.84	& 12.43	& 11.65	& 11.25		& 9.34		& 7.80	& 6.32	\\
\rowcolor{LightCyan}
SLEB		& T. Block & 20\% & 381 & 1.27$\times$ & 15.99	& 13.81	& 12.74	& 12.54		& 12.32	& 9.42	& 7.31	\\
\bottomrule
\end{tabular}
\end{table*}

\subsection{Language Modeling}

We evaluate the linguistic capabilities of pruned LLMs on language modeling tasks.
Table~\ref{table:c4_ppl} showcases the perplexity results on C4 validation dataset~\cite{c4-t5}.
It is important to note that though the sparsity of SLEB is 10\% to 20\% whereas previous works show a sparsity ratio between 20\% to 50\%, as we will discuss in Section~\ref{sec:speedup}, SLEB outperforms in LLM inference speedup.
Though Wanda and DSnoT completely fail on OPT-66B, SLEB sufficiently preserves perplexity scores for all evaluated cases.
Moreover, SLEB outperforms previous pruning methods in the majority of cases by a significant margin, though it uses a coarser granularity of pruning unit.
These results once again emphasize a point discussed in Section~\ref{sec:similarity} - LLMs exhibit robustness at the level of transformer blocks. By effectively identifying and removing transformer blocks, SLEB can have a minimal impact on perplexity scores.
The perplexity results on WikiText-2 test dataset can be found in Appendix~\ref{appendix:more_language_modeling}.

\subsection{Dependency on Calibration Dataset}
\label{subsec:dependency}

We analyze the influence of the calibration data on the perplexity results of pruned LLMs.
During the pruning process of the target LLMs, 128 sequences are randomly sampled either from WikiText-2 or C4 training datasets to serve as the calibration data.
Subsequently, the pruned LLMs are evaluated using both WikiText-2 and C4 datasets.
We compare pruning results of SparseGPT, Wanda, SliceGPT with 25\% sparsity, and SLEB with 20\% sparsity.

\begin{figure}[t]
    \begin{subfigure}{\columnwidth}
      \centering
      \includegraphics[width=\columnwidth]{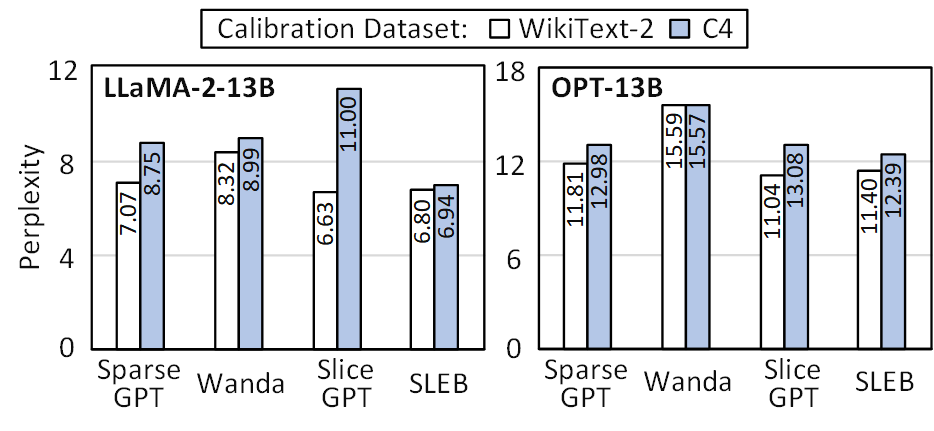}
      \vspace{-6mm}
      \caption{Perplexity results on WikiText-2}
    \end{subfigure}
    \begin{subfigure}{\columnwidth}
      \centering
      \includegraphics[width=\columnwidth]{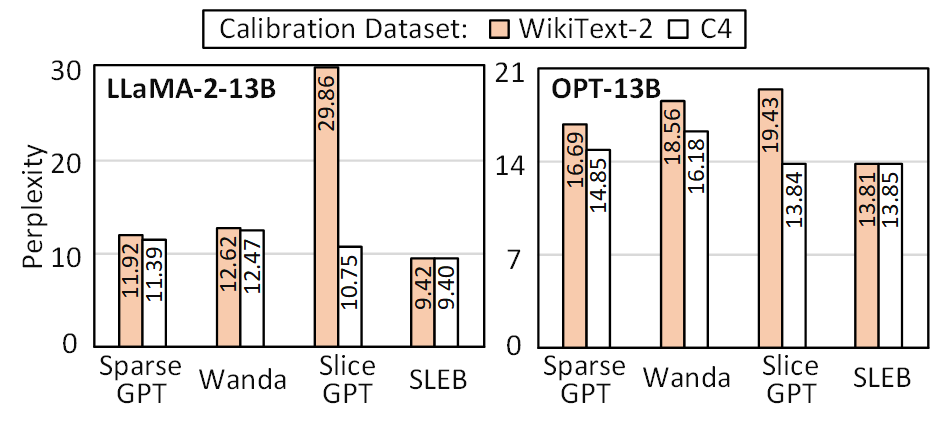}
      \vspace{-6mm}
      \caption{Perplexity results on C4}
    \end{subfigure}
    \vspace{-7mm}
    \caption{Perplexity results of pruned LLMs depend on the calibration dataset. 
    The colored bar graphs depict the results when there is a difference between the calibration dataset and the evaluation dataset.
    }
    \label{fig:data_dependency}
    \vspace{-10mm}
\end{figure}

\begin{table*}[t]
\caption{
Mean accuracies (\%) on zero-shot tasks and latency results. We measure latency of each method with LLaMA-2-70B on 2 NVIDIA A100 GPUs. (T.Block: Transformer Block)
}
\label{table:zeroshot_avg}
 \centering
  \footnotesize
\begin{tabular}{l | M{1.2cm} |  c | c | c | cccc | ccc}
\toprule
\rowcolor{Gray}
 ~ & \bf Pruning & ~ & ~ & ~ & \multicolumn{4}{c|}{\bf OPT} & \multicolumn{3}{c}{\bf LLaMA-2}\\
\rowcolor{Gray}
\bf Method & \bf Unit & \bf Sparsity &  \bf Latency(ms) &  \bf Speedup &  \bf 6.7B & \bf 13B & \bf 30B & \bf 66B & \bf 7B & \bf 13B & \bf 70B \\
\midrule
\midrule
Dense		& - & 0\%  & 1718.4 & 1.00$\times$ & 60.70	& 61.79	& 64.40	& 66.16 & 69.00	& 71.76	& 76.57 \\
\midrule
SparseGPT	& 2:4 & 50\%  & 1555.5 & 1.10$\times$ & 54.94	& 56.76	& 59.96	& 62.33 & 58.23	& 63.06	& 71.87 \\
Wanda		& 2:4 & 50\%  & 1555.5 & 1.10$\times$ & 53.14	& 55.12	& 58.89	& 35.93 & 55.59	& 61.23	& 72.34 \\
\midrule
SliceGPT	& Channel & 20\% & 1658.7 & 1.04$\times$ & 56.31	& 60.20	& 63.65	& 65.74 & 58.17	& 63.45	& 72.34 \\
SliceGPT	& Channel & 25\% & 1440.7 & 1.19$\times$ & 54.28	& 59.27	& 62.11	& 65.17 & 55.49	& 58.90	& 69.75 \\
SliceGPT	& Channel & 30\% & 1364.2 & 1.26$\times$ & 53.00	& 57.42	& 61.27	& 64.24 & 51.50	& 55.16	& 66.11 \\
\midrule
\rowcolor{LightCyan}
SLEB		& T. Block & 10\% & 1529.1 & 1.12$\times$ & 60.00	& 62.07	& 64.48	& 65.38 & 62.24	& 66.77	& 73.14 \\
\rowcolor{LightCyan}
SLEB		& T. Block & 20\% & 1364.1 & 1.26$\times$ & 57.61	& 60.08	& 62.86	& 62.53 & 56.80	& 62.96	& 70.81 \\

\bottomrule
\end{tabular}
\vspace{-3mm}
\end{table*}

As shown in Figure~\ref{fig:data_dependency}, SLEB demonstrates the least dependency on the choice of the calibration dataset.
SliceGPT struggles to preserve perplexity scores when the type of evaluation data differs from the type of calibration data.
SliceGPT relies on the principal component analysis of activation matrices to determine which neurons to prune for channel-wise pruning, making its performance more reliant on the calibration dataset.
SparseGPT and Wanda take a different approach by measuring the significance of weights in the matrix multiplication results of each linear layer. These methods leverage the information contained in the pre-trained weight matrices of LLMs and show less dependency on the dataset.
While previous methods measure redundancy at the layer level, SLEB assesses the redundancy of each transformer block at the entire network level using $Metric^3$. This approach fully exploits the information present in the pre-trained LLMs and demonstrates less dependency on the dataset.
More results comparing the dependency on the calibration dataset are presented in Appendix ~\ref{appendix:more_dependency}.


\subsection{Zero-shot Tasks}

Following the evaluation methods presented in previous works~\cite{slicegpt}, we evaluate SLEB on following zero-shot tasks: PIQA~\cite{piqa}, WinoGrande~\cite{winogrande}, HellaSwag~\cite{hellaswag}, ARC-easy and ARC-challenge~\cite{arc}.
LM Evaluation Harness~\cite{eval-harness} with its default parameters is used for the evaluation.

Table~\ref{table:zeroshot_avg} shows the average accuracies of pruned LLMs on zero-shot tasks.
SLEB with 10\% sparsity achieves superior accuracies compared to previous pruning methods, preserving the original accuracies to a significant extent.
Even when the sparsity of SLEB is increased to 20\%, it continues to outperform previous approaches in most cases.
The task-wise accuracies are provided in Appendix~\ref{appendix:more_zeroshot}.

\subsection{Speedup}
\label{sec:speedup}

We evaluate the LLM inference speedup achieved by SLEB in comparison to previous pruning methods - 2:4 pruning and 25\% channel-wise pruning.
The inference speed is assessed using the LLM implementation provided by HuggingFace Transformers libraries for each pruning method.
The language processing of LLMs can be categorized into two stages: prompt processing and token generation.
In the prompt processing stage, LLMs assess the given prompt and generate the KV cache.
In the token generation stage, LLMs generate new tokens in an auto-regressive manner.
Because these two stages have distinct characteristics in terms of inference speed, with the prompt processing stage tending to be compute-bound and the token generation stage tending to be memory-bound, we analyze the speedup in each of them separately.

Table~\ref{table:c4_ppl} and Table~\ref{table:zeroshot_avg} present throughput and latency results for LLaMA-2-70B with 2 NVIDIA A100 GPUs during token generation and prompt processing, respectively.
For token generation, the test scenario consists of generating sentences with a length of 128 tokens and a batch size of 64.
For prompt processing, we measure the latency when processing an input sequence with 2048 tokens. 
In both scenarios, SLEB demonstrates improvements in latency and throughput that are proportional to the pruning ratio. This performance enhancement is attributed to the adoption of transformer block, which is a fundamental building block of LLMs, as the basic unit of pruning.
In contrast, previous approaches fail to achieve significant improvements, particularly in the case of token generation.
For more detailed analysis results, please refer to Appendix~\ref{appendix:more_speedup}.
SLEB consistently provides speedup benefits across various serving scenarios, while the speedup achieved by other methods is significantly influenced by the specific serving scenarios.

\subsection{Compatibility with Post-Training Quantization}

\begin{figure}[t]
      \centering
      \vspace{1mm}
      \includegraphics[width=0.9\columnwidth]{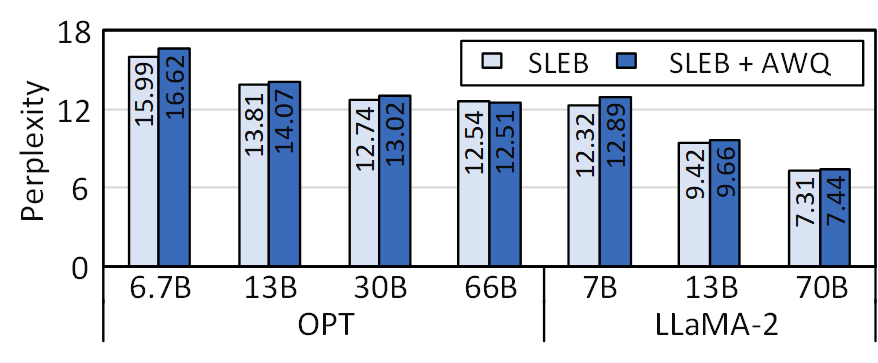}
      \vspace{-6mm}
    \caption{Perplexity comparison between LLMs pruned with SLEB (target sparsity: 20\%) and those further compressed using AWQ, a 4-bit weight quantization}
    \label{fig:sleb_ptq_c4}
    \vspace{-8mm}
\end{figure}


Post-training quantization (PTQ) is a well-established technique for compressing and accelerating LLMs~\cite{gptq, owq, awq}. 
If PTQ can be applied to compress LLMs alongside SLEB, we can expect further improvements in memory efficiency and LLM inference speed. 
In Figure~\ref{fig:sleb_ptq_c4}, we present a comparison of perplexity scores for LLMs that have undergone SLEB with target sparsity of 20\% and those that have been further compressed with AWQ~\cite{awq}, a state-of-the-art PTQ algorithm that quantizes weights to 4-bit integer values.
The experimental results indicate that applying PTQ to streamlined models has no discernible impact on their perplexity.
More analysis results on compatibility with PTQ are provided in Appendix~\ref{appendix:ptq}.

\section{Conclusion}
\label{sec:conclusion}

This paper introduces SLEB, a novel technique for streamlining LLMs by removing redundant transformer blocks.
SLEB carefully identifies and eliminates redundant blocks, ensuring that it effectively preserves the language capabilities of LLMs without the need for any resource-intensive training.
Our experimental results demonstrate that SLEB can successfully prune up to 20\% of blocks while maintaining the linguistic performance of LLMs across language modeling and zero-shot tasks.
One of the significant advantages of SLEB is its ability to remove entire transformer blocks, resulting in substantial speed improvements in end-to-end LLM inference. This speedup is applicable across various implementation scenarios, making SLEB a practical solution for real-world LLM serving scenarios.


\section*{Acknowledgements}
This work was supported in part by Institute of Information \& communications Technology Planning \& Evaluation (IITP) grant funded by the Korea government (MSIT) (No. 2021-0-00105: Development of model compression framework for scalable on-device AI computing on Edge applications , IITP-2023-RS-2023-00256081: artificial intelligence semiconductor support program to nurture the best talents, No. 2021-0-01343: Artificial Intelligence Graduate School Program (Seoul National University) , No.2021-0-02068: Artificial Intelligence Innovation Hub), in part by the Samsung Research Funding Center under Project SRFC-TC1603-53, and BK21 FOUR program at Seoul National University.



\section*{Impact Statement}
This paper presents work whose goal is to advance the field of Machine Learning. There are many potential societal consequences of our work, none which we feel must be specifically highlighted here.

\bibliography{icml}

\begin{thebibliography}{37}
\providecommand{\natexlab}[1]{#1}
\providecommand{\url}[1]{\texttt{#1}}
\expandafter\ifx\csname urlstyle\endcsname\relax
  \providecommand{\doi}[1]{doi: #1}\else
  \providecommand{\doi}{doi: \begingroup \urlstyle{rm}\Url}\fi

\bibitem[Ashkboos et~al.(2024)Ashkboos, Croci, do~Nascimento, Hoefler, and Hensman]{slicegpt}
Ashkboos, S., Croci, M.~L., do~Nascimento, M.~G., Hoefler, T., and Hensman, J.
\newblock Slice{GPT}: Compress large language models by deleting rows and columns.
\newblock In \emph{The Twelfth International Conference on Learning Representations}, 2024.

\bibitem[Bisk et~al.(2020)Bisk, Zellers, Le~bras, Gao, and Choi]{piqa}
Bisk, Y., Zellers, R., Le~bras, R., Gao, J., and Choi, Y.
\newblock Piqa: Reasoning about physical commonsense in natural language.
\newblock \emph{Proceedings of the AAAI Conference on Artificial Intelligence}, 34\penalty0 (05):\penalty0 7432--7439, Apr. 2020.
\newblock \doi{10.1609/aaai.v34i05.6239}.
\newblock URL \url{https://ojs.aaai.org/index.php/AAAI/article/view/6239}.

\bibitem[Brown et~al.(2020)Brown, Mann, Ryder, Subbiah, Kaplan, Dhariwal, Neelakantan, Shyam, Sastry, Askell, et~al.]{gpt3}
Brown, T., Mann, B., Ryder, N., Subbiah, M., Kaplan, J.~D., Dhariwal, P., Neelakantan, A., Shyam, P., Sastry, G., Askell, A., et~al.
\newblock Language models are few-shot learners.
\newblock \emph{Advances in neural information processing systems}, 33:\penalty0 1877--1901, 2020.

\bibitem[Chen et~al.(2023)Chen, Pan, Li, Ding, and Zhou]{ee_llm}
Chen, Y., Pan, X., Li, Y., Ding, B., and Zhou, J.
\newblock Ee-llm: Large-scale training and inference of early-exit large language models with 3d parallelism.
\newblock \emph{arXiv preprint arXiv:2312.04916}, 2023.

\bibitem[Chowdhery et~al.(2022)Chowdhery, Narang, Devlin, Bosma, Mishra, Roberts, Barham, Chung, Sutton, Gehrmann, Schuh, Shi, Tsvyashchenko, Maynez, Rao, Barnes, Tay, Shazeer, Prabhakaran, Reif, Du, Hutchinson, Pope, Bradbury, Austin, Isard, Gur-Ari, Yin, Duke, Levskaya, Ghemawat, Dev, Michalewski, Garcia, Misra, Robinson, Fedus, Zhou, Ippolito, Luan, Lim, Zoph, Spiridonov, Sepassi, Dohan, Agrawal, Omernick, Dai, Pillai, Pellat, Lewkowycz, Moreira, Child, Polozov, Lee, Zhou, Wang, Saeta, Diaz, Firat, Catasta, Wei, Meier-Hellstern, Eck, Dean, Petrov, and Fiedel]{palm}
Chowdhery, A., Narang, S., Devlin, J., Bosma, M., Mishra, G., Roberts, A., Barham, P., Chung, H.~W., Sutton, C., Gehrmann, S., Schuh, P., Shi, K., Tsvyashchenko, S., Maynez, J., Rao, A., Barnes, P., Tay, Y., Shazeer, N., Prabhakaran, V., Reif, E., Du, N., Hutchinson, B., Pope, R., Bradbury, J., Austin, J., Isard, M., Gur-Ari, G., Yin, P., Duke, T., Levskaya, A., Ghemawat, S., Dev, S., Michalewski, H., Garcia, X., Misra, V., Robinson, K., Fedus, L., Zhou, D., Ippolito, D., Luan, D., Lim, H., Zoph, B., Spiridonov, A., Sepassi, R., Dohan, D., Agrawal, S., Omernick, M., Dai, A.~M., Pillai, T.~S., Pellat, M., Lewkowycz, A., Moreira, E., Child, R., Polozov, O., Lee, K., Zhou, Z., Wang, X., Saeta, B., Diaz, M., Firat, O., Catasta, M., Wei, J., Meier-Hellstern, K., Eck, D., Dean, J., Petrov, S., and Fiedel, N.
\newblock Palm: Scaling language modeling with pathways, 2022.

\bibitem[Clark et~al.(2018)Clark, Cowhey, Etzioni, Khot, Sabharwal, Schoenick, and Tafjord]{arc}
Clark, P., Cowhey, I., Etzioni, O., Khot, T., Sabharwal, A., Schoenick, C., and Tafjord, O.
\newblock Think you have solved question answering? try arc, the ai2 reasoning challenge, 2018.

\bibitem[Del~Corro et~al.(2023)Del~Corro, Del~Giorno, Agarwal, Yu, Awadallah, and Mukherjee]{skipdecode}
Del~Corro, L., Del~Giorno, A., Agarwal, S., Yu, B., Awadallah, A., and Mukherjee, S.
\newblock Skipdecode: Autoregressive skip decoding with batching and caching for efficient llm inference.
\newblock \emph{arXiv preprint arXiv:2307.02628}, 2023.

\bibitem[Din et~al.(2023)Din, Karidi, Choshen, and Geva]{jump_to_con}
Din, A.~Y., Karidi, T., Choshen, L., and Geva, M.
\newblock Jump to conclusions: Short-cutting transformers with linear transformations.
\newblock \emph{arXiv preprint arXiv:2303.09435}, 2023.

\bibitem[Frantar \& Alistarh(2022)Frantar and Alistarh]{obc}
Frantar, E. and Alistarh, D.
\newblock Optimal brain compression: A framework for accurate post-training quantization and pruning.
\newblock \emph{Advances in Neural Information Processing Systems}, 35:\penalty0 4475--4488, 2022.

\bibitem[Frantar \& Alistarh(2023)Frantar and Alistarh]{sparsegpt}
Frantar, E. and Alistarh, D.
\newblock Sparsegpt: Massive language models can be accurately pruned in one-shot.
\newblock In \emph{International Conference on Machine Learning}, pp.\  10323--10337. PMLR, 2023.

\bibitem[Frantar et~al.(2023)Frantar, Ashkboos, Hoefler, and Alistarh]{gptq}
Frantar, E., Ashkboos, S., Hoefler, T., and Alistarh, D.
\newblock Gptq: Accurate post-training quantization for generative pre-trained transformers.
\newblock \emph{International Conference on Learning Representations}, 2023.

\bibitem[Gao et~al.(2023)Gao, Tow, Abbasi, Biderman, Black, DiPofi, Foster, Golding, Hsu, Le~Noac'h, Li, McDonell, Muennighoff, Ociepa, Phang, Reynolds, Schoelkopf, Skowron, Sutawika, Tang, Thite, Wang, Wang, and Zou]{eval-harness}
Gao, L., Tow, J., Abbasi, B., Biderman, S., Black, S., DiPofi, A., Foster, C., Golding, L., Hsu, J., Le~Noac'h, A., Li, H., McDonell, K., Muennighoff, N., Ociepa, C., Phang, J., Reynolds, L., Schoelkopf, H., Skowron, A., Sutawika, L., Tang, E., Thite, A., Wang, B., Wang, K., and Zou, A.
\newblock A framework for few-shot language model evaluation, 12 2023.
\newblock URL \url{https://zenodo.org/records/10256836}.

\bibitem[Han et~al.(2015)Han, Pool, Tran, and Dally]{songhan}
Han, S., Pool, J., Tran, J., and Dally, W.
\newblock Learning both weights and connections for efficient neural network.
\newblock \emph{Advances in neural information processing systems}, 28, 2015.

\bibitem[Hassibi et~al.(1993)Hassibi, Stork, and Wolff]{obs}
Hassibi, B., Stork, D.~G., and Wolff, G.~J.
\newblock Optimal brain surgeon and general network pruning.
\newblock In \emph{IEEE international conference on neural networks}, pp.\  293--299. IEEE, 1993.

\bibitem[Hu et~al.(2021)Hu, Wallis, Allen-Zhu, Li, Wang, Wang, Chen, et~al.]{lora}
Hu, E.~J., Wallis, P., Allen-Zhu, Z., Li, Y., Wang, S., Wang, L., Chen, W., et~al.
\newblock Lora: Low-rank adaptation of large language models.
\newblock In \emph{International Conference on Learning Representations}, 2021.

\bibitem[LeCun et~al.(1989)LeCun, Denker, and Solla]{obd}
LeCun, Y., Denker, J., and Solla, S.
\newblock Optimal brain damage.
\newblock \emph{Advances in neural information processing systems}, 2, 1989.

\bibitem[Lee et~al.(2024)Lee, Jin, Kim, Kim, and Park]{owq}
Lee, C., Jin, J., Kim, T., Kim, H., and Park, E.
\newblock Owq: Lessons learned from activation outliers for weight quantization in large language models.
\newblock In \emph{Proceedings of the AAAI Conference on Artificial Intelligence}, 2024.

\bibitem[Lin et~al.(2023)Lin, Tang, Tang, Yang, Dang, and Han]{awq}
Lin, J., Tang, J., Tang, H., Yang, S., Dang, X., and Han, S.
\newblock Awq: Activation-aware weight quantization for llm compression and acceleration.
\newblock \emph{arXiv preprint arXiv:2306.00978}, 2023.

\bibitem[Liu et~al.(2023)Liu, Wang, Dao, Zhou, Yuan, Song, Shrivastava, Zhang, Tian, Re, et~al.]{dejavu}
Liu, Z., Wang, J., Dao, T., Zhou, T., Yuan, B., Song, Z., Shrivastava, A., Zhang, C., Tian, Y., Re, C., et~al.
\newblock Deja vu: Contextual sparsity for efficient llms at inference time.
\newblock In \emph{International Conference on Machine Learning}, pp.\  22137--22176. PMLR, 2023.

\bibitem[Ma et~al.(2023)Ma, Fang, and Wang]{llm_pruner}
Ma, X., Fang, G., and Wang, X.
\newblock Llm-pruner: On the structural pruning of large language models.
\newblock In \emph{Thirty-seventh Conference on Neural Information Processing Systems}, 2023.

\bibitem[Merity et~al.(2016)Merity, Xiong, Bradbury, and Socher]{wikitext2}
Merity, S., Xiong, C., Bradbury, J., and Socher, R.
\newblock Pointer sentinel mixture models.
\newblock \emph{arXiv preprint arXiv:1609.07843}, 2016.

\bibitem[Mishra et~al.(2021)Mishra, Latorre, Pool, Stosic, Stosic, Venkatesh, Yu, and Micikevicius]{24gpu}
Mishra, A., Latorre, J.~A., Pool, J., Stosic, D., Stosic, D., Venkatesh, G., Yu, C., and Micikevicius, P.
\newblock Accelerating sparse deep neural networks.
\newblock \emph{arXiv preprint arXiv:2104.08378}, 2021.

\bibitem[Paszke et~al.(2019)Paszke, Gross, Massa, Lerer, Bradbury, Chanan, Killeen, Lin, Gimelshein, Antiga, et~al.]{pytorch}
Paszke, A., Gross, S., Massa, F., Lerer, A., Bradbury, J., Chanan, G., Killeen, T., Lin, Z., Gimelshein, N., Antiga, L., et~al.
\newblock Pytorch: An imperative style, high-performance deep learning library.
\newblock \emph{Advances in neural information processing systems}, 32, 2019.

\bibitem[Raffel et~al.(2020)Raffel, Shazeer, Roberts, Lee, Narang, Matena, Zhou, Li, and Liu]{c4-t5}
Raffel, C., Shazeer, N., Roberts, A., Lee, K., Narang, S., Matena, M., Zhou, Y., Li, W., and Liu, P.~J.
\newblock Exploring the limits of transfer learning with a unified text-to-text transformer.
\newblock \emph{Journal of Machine Learning Research}, 21\penalty0 (140):\penalty0 1--67, 2020.
\newblock URL \url{http://jmlr.org/papers/v21/20-074.html}.

\bibitem[Sakaguchi et~al.(2019)Sakaguchi, Bras, Bhagavatula, and Choi]{winogrande}
Sakaguchi, K., Bras, R.~L., Bhagavatula, C., and Choi, Y.
\newblock Winogrande: An adversarial winograd schema challenge at scale.
\newblock \emph{arXiv preprint arXiv:1907.10641}, 2019.

\bibitem[Schuster et~al.(2022)Schuster, Fisch, Gupta, Dehghani, Bahri, Tran, Tay, and Metzler]{calm}
Schuster, T., Fisch, A., Gupta, J., Dehghani, M., Bahri, D., Tran, V., Tay, Y., and Metzler, D.
\newblock Confident adaptive language modeling.
\newblock \emph{Advances in Neural Information Processing Systems}, 35:\penalty0 17456--17472, 2022.

\bibitem[Shi et~al.(2020)Shi, Wang, and Chu]{sparsegpu}
Shi, S., Wang, Q., and Chu, X.
\newblock Efficient sparse-dense matrix-matrix multiplication on gpus using the customized sparse storage format.
\newblock In \emph{2020 IEEE 26th International Conference on Parallel and Distributed Systems (ICPADS)}, pp.\  19--26. IEEE, 2020.

\bibitem[Sun et~al.(2023)Sun, Liu, Bair, and Kolter]{wanda}
Sun, M., Liu, Z., Bair, A., and Kolter, J.~Z.
\newblock A simple and effective pruning approach for large language models.
\newblock \emph{arXiv preprint arXiv:2306.11695}, 2023.

\bibitem[Taori et~al.(2023)Taori, Gulrajani, Zhang, Dubois, Li, Guestrin, Liang, and Hashimoto]{alpaca}
Taori, R., Gulrajani, I., Zhang, T., Dubois, Y., Li, X., Guestrin, C., Liang, P., and Hashimoto, T.~B.
\newblock Stanford alpaca: An instruction-following llama model.
\newblock \url{https://github.com/tatsu-lab/stanford_alpaca}, 2023.

\bibitem[Touvron et~al.(2023{\natexlab{a}})Touvron, Lavril, Izacard, Martinet, Lachaux, Lacroix, Rozière, Goyal, Hambro, Azhar, Rodriguez, Joulin, Grave, and Lample]{llama_v1}
Touvron, H., Lavril, T., Izacard, G., Martinet, X., Lachaux, M.-A., Lacroix, T., Rozière, B., Goyal, N., Hambro, E., Azhar, F., Rodriguez, A., Joulin, A., Grave, E., and Lample, G.
\newblock Llama: Open and efficient foundation language models, 2023{\natexlab{a}}.

\bibitem[Touvron et~al.(2023{\natexlab{b}})Touvron, Martin, Stone, Albert, Almahairi, Babaei, Bashlykov, Batra, Bhargava, Bhosale, et~al.]{llama_v2}
Touvron, H., Martin, L., Stone, K., Albert, P., Almahairi, A., Babaei, Y., Bashlykov, N., Batra, S., Bhargava, P., Bhosale, S., et~al.
\newblock Llama 2: Open foundation and fine-tuned chat models.
\newblock \emph{arXiv preprint arXiv:2307.09288}, 2023{\natexlab{b}}.

\bibitem[Varshney et~al.(2023)Varshney, Chatterjee, Parmar, and Baral]{lite}
Varshney, N., Chatterjee, A., Parmar, M., and Baral, C.
\newblock Accelerating llama inference by enabling intermediate layer decoding via instruction tuning with lite.
\newblock \emph{arXiv e-prints}, pp.\  arXiv--2310, 2023.

\bibitem[Wang(2020)]{sparsert}
Wang, Z.
\newblock Sparsert: Accelerating unstructured sparsity on gpus for deep learning inference.
\newblock In \emph{Proceedings of the ACM international conference on parallel architectures and compilation techniques}, pp.\  31--42, 2020.

\bibitem[Wolf et~al.(2020)Wolf, Debut, Sanh, Chaumond, Delangue, Moi, Cistac, Rault, Louf, Funtowicz, Davison, Shleifer, von Platen, Ma, Jernite, Plu, Xu, Scao, Gugger, Drame, Lhoest, and Rush]{huggingface-transformers}
Wolf, T., Debut, L., Sanh, V., Chaumond, J., Delangue, C., Moi, A., Cistac, P., Rault, T., Louf, R., Funtowicz, M., Davison, J., Shleifer, S., von Platen, P., Ma, C., Jernite, Y., Plu, J., Xu, C., Scao, T.~L., Gugger, S., Drame, M., Lhoest, Q., and Rush, A.~M.
\newblock Transformers: State-of-the-art natural language processing.
\newblock In \emph{Proceedings of the 2020 Conference on Empirical Methods in Natural Language Processing: System Demonstrations}, pp.\  38--45, Online, October 2020. Association for Computational Linguistics.
\newblock URL \url{https://www.aclweb.org/anthology/2020.emnlp-demos.6}.

\bibitem[Zellers et~al.(2019)Zellers, Holtzman, Bisk, Farhadi, and Choi]{hellaswag}
Zellers, R., Holtzman, A., Bisk, Y., Farhadi, A., and Choi, Y.
\newblock Hellaswag: Can a machine really finish your sentence?
\newblock In \emph{Proceedings of the 57th Annual Meeting of the Association for Computational Linguistics}, 2019.

\bibitem[Zhang et~al.(2022)Zhang, Roller, Goyal, Artetxe, Chen, Chen, Dewan, Diab, Li, Lin, Mihaylov, Ott, Shleifer, Shuster, Simig, Koura, Sridhar, Wang, and Zettlemoyer]{opt}
Zhang, S., Roller, S., Goyal, N., Artetxe, M., Chen, M., Chen, S., Dewan, C., Diab, M., Li, X., Lin, X.~V., Mihaylov, T., Ott, M., Shleifer, S., Shuster, K., Simig, D., Koura, P.~S., Sridhar, A., Wang, T., and Zettlemoyer, L.
\newblock Opt: Open pre-trained transformer language models, 2022.

\bibitem[Zhang et~al.(2024)Zhang, Zhao, Lin, Sun, Yao, Han, Tanner, Liu, and Ji]{DSnoT}
Zhang, Y., Zhao, L., Lin, M., Sun, Y., Yao, Y., Han, X., Tanner, J., Liu, S., and Ji, R.
\newblock Dynamic sparse no training: Training-free fine-tuning for sparse llms.
\newblock In \emph{The Twelfth International Conference on Learning Representations}, 2024.

\end{thebibliography}
\bibliographystyle{icml2024}

\newpage
\appendix
\onecolumn


\newpage
\section{SLEB details}
\label{appendix:SLEB_details}

\subsection{Runtime}
\label{appendix:runtime}

We compare runtime of SLEB and previous works for pruning LLaMA-2 models. Here, the runtime indicates the time required to execute the pruning algorithms. Specifically, it denotes the time needed to identify and remove elements based on each particular pruning approach. It is important to note that each pruning algorithm is applied to each model only once. Therefore, as long as the runtime remains within a reasonable range, the application of the pruning algorithms can proceed smoothly without any complications.

The official implementation of SparseGPT~\footnote{https://github.com/IST-DASLab/sparsegpt} offloads the entire LLMs to the CPU and loads the target transformer block to GPU one by one for pruning, resulting in long runtime.
Hence, for both SparseGPT and Wanda, we utilize the official implementation of Wanda~\footnote{https://github.com/locuslab/wanda}, which supports distributing LLMs across multiple GPUs.
For SliceGPT, we use its official implementation~\footnote{https://github.com/microsoft/TransformerCompression/tree/main}.
As shown in Table~\ref{table:runtime_comparison}, runtime of SLEB is generally within the range of runtime of other pruning methods.


\begin{table}[h]
\centering
\footnotesize
\caption{Comparison of runtime for pruning LLaMA-2 using NVIDIA A100 GPU}
\label{table:runtime_comparison}
\begin{tabular}{l | cc | cc | cc}
\toprule
 ~ & \multicolumn{2}{c|}{7B} & \multicolumn{2}{c|}{13B} & \multicolumn{2}{c}{70B} \\
  Method & \#GPU & Time (s) & \#GPU & Time (s) & \#GPU & Time (s) \\ \midrule
SparseGPT & 1 & 528 & 1 & 920 & 3 & 5040 \\
Wanda & 1 & 64 & 1 & 104 & 2 & 480 \\
SliceGPT & 1 & 452 & 1 & 633 & 1 & 3010 \\
SLEB 20\% & 1 & 113 & 1 & 279 & 2 & 5420 \\
\bottomrule
\end{tabular}
\end{table}


\subsection{Selected Transformer Blocks}
\label{appendix:selected_blocks}

We summarize the number and indices of transformer blocks selected for removal using SLEB in Table~\ref{table:idx_removed blocks}. 
The order of indices presented in Table~\ref{table:idx_removed blocks} corresponds to the order in which they are removed.

\begin{table}[h]
\caption{Number and indices of removed blocks when applying SLEB on different LLM models. In 20\% sparsity scenarios, it initially removes blocks selected for 10\% sparsity and then selects additional blocks for removal to reach the desired sparsity level.}
\label{table:idx_removed blocks}
 \centering
  \footnotesize
\begin{tabular}{l  l  c | c | l | c | l }
\toprule

\rowcolor{Gray}
\multicolumn{3}{c|}{\bf Model} & \multicolumn{2}{c|}{\bf  10\% Sparsity} & \multicolumn{2}{c}{\bf 20\% Sparsity} \\
\rowcolor{Gray}
\bf Type & \bf Size & \bf \# Blocks &	\bf \# Removed	&	\multicolumn{1}{c|}{\bf Removed Indices}	&	\bf \# Removed	& \multicolumn{1}{c}{\bf Removed Indices}	\\
\midrule
\midrule
\multirow{4}{*}{\hspace{-0.15cm}\begin{tabular}{c}\\[-8pt]\text{OPT}\end{tabular}}	&	6.7B	&	32	&	4	&	6, 7, 3, 24	&	7	&	+ 18, 30, 11	\\
	&	13B	&	40	&	4	&	5, 4, 9, 2	&	8	&	+ 14, 25, 34, 10	\\
	&	30B	&	48	&	5	&	4, 17, 7, 12, 14	&	10	&	+ 40, 23, 2, 13, 43	\\
	&	66B	&	64	&	7	&	18, 12, 21, 46, 8, 16, 9	&	13	&	+ 65, 13, 27, 61, 35, 42	\\
 \midrule
\multirow{3}{*}{\hspace{-0.15cm}\begin{tabular}{c}\\[-8pt]\text{LLaMA-2}\end{tabular}}	&	7B	&	32	&	4	&	14, 23, 11, 24	&	7	&	+ 10, 27, 15	\\
	&	13B	&	40	&	4	&	31, 27, 13, 12	&	8	&	+ 33, 29, 14, 2	\\
	&	70B	&	80	&	8	&	33, 29, 61, 11, 55, 30, 59, 28	&	16	&	+ 50, 27, 68, 32, 31, 15, 60, 4	\\

\bottomrule


\end{tabular}
\end{table}

\newpage
\section{More Evaluation Results}
\label{appendix:more_eval_results}

\subsection{Limitations of Early Exit}
\label{appendix:lim_of_early_exit}

We analyze the ratio of skipped blocks in the LLMs under the assumption of an exit at the ideal point.
The ideal exit point is defined as the point at which the token prediction outcomes of a specific transformer block align with the final predictions of the LLMs. 
The results are depicted in Table ~\ref{tab:ee_prune_ratio}.

\begin{table}[h]
\centering
\footnotesize
\caption{The ratio of skipped blocks under the assumption of early exiting at ideal exit points}
\label{tab:ee_prune_ratio}
\begin{tabular}{c|c|c|c|c|c|c}
\toprule
\rowcolor{Gray}
Batch & LLaMA-2-7B & LLaMA-2-13B & LLaMA-2-70B & OPT-6.7B & OPT-13B & OPT-66B \\ \midrule
1 & 8.23\% & 9.56\% & 8.96\% & 12.52\% & 11.74\% & 11.89\% \\ \bottomrule
\end{tabular}
\end{table}

Next, we evaluate the throughput improvement achieved by implementing ideal early exits in scenarios where sentences of 128 tokens are generated across different batch sizes. These results are presented in Table ~\ref{tab:ee_speedup}.
In the single-batch scenario, the speedup of early exit is directly proportional to the ratio of skipped blocks. Consequently, early exit can achieve up to a 1.14$\times$ speedup. 
Conversely, in multi-batch scenarios, where multiple tokens are processed simultaneously, the inference of the LLMs to follow the longest path dictated by any of the tokens. 
For instance, if in a batch of four tokens, three tokens can complete processing halfway through the LLM while the fourth requires full traversal, the entire batch must process through the full length of the LLM to accommodate the token with the longest path. 
Therefore, as the batch size increases, the effectiveness of early exit diminishes since the termination point of each token differs, reducing the likelihood of achieving a significant speedup.

\begin{table}[h]
\centering
\footnotesize
\caption{Throughput improvement under the assumption of early exiting at ideal exit points}
\label{tab:ee_speedup}
\begin{tabular}{c|c|c|c|c|c|c}
\toprule
\rowcolor{Gray}
Batch & LLaMA-2-7B & LLaMA-2-13B & LLaMA-2-70B & OPT-6.7B & OPT-13B & OPT-66B \\ \midrule
1 & 1.09$\times$ & 1.11$\times$ & 1.10$\times$ & 1.14$\times$ & 1.13$\times$ & 1.13$\times$ \\ \midrule
2 & 1.03$\times$ & 1.02$\times$ & 1.02$\times$ & 1.05$\times$ & 1.06$\times$ & 1.05$\times$ \\ \midrule
4 & 1.00$\times$ & 1.00$\times$ & 1.00$\times$ & 1.02$\times$ & 1.02$\times$ & 1.02$\times$ \\ \midrule
8 & 1.00$\times$ & 1.00$\times$ & 1.00$\times$ & 1.00$\times$ & 1.01$\times$ & 1.00$\times$ \\ \midrule
16 & 1.00$\times$ & 1.00$\times$ & 1.00$\times$ & 1.00$\times$ & 1.00$\times$ & 1.00$\times$ \\ \midrule
32 & 1.00$\times$ & 1.00$\times$ & 1.00$\times$ & 1.00$\times$ & 1.00$\times$ & 1.00$\times$ \\ \midrule
64 & 1.00$\times$ & 1.00$\times$ & 1.00$\times$ & 1.00$\times$ & 1.00$\times$ & 1.00$\times$ \\ \bottomrule
\end{tabular}
\end{table}

\newpage
\subsection{Language Modeling}
\label{appendix:more_language_modeling}

We use sequence length 2048 for all models to evaluate perplexity.
Table~\ref{table:wiki_ppl} provides perplexity results of pruned models on WikiText-2 dataset.
Notably, SliceGPT demonstrates good perplexity in this case because it uses calibration data sampled from the WikiText-2 training dataset.
As discussed in Section~\ref{subsec:dependency}, SliceGPT tends to perform well in terms of perplexity when the types of calibration and evaluation datasets align. However, it may yield poor results when the datasets do not align. This behavior is attributed to SliceGPT relying solely on information from activation matrices to determine the target channels for pruning.
Hence, when SliceGPT is evaluated on C4 dataset, it shows poor perplexity results (Table~\ref{table:c4_ppl}).

\begin{table}[h]
\caption{
Perplexity results on WikiText-2 dataset and throughput (tokens/s) results. We measure throughput of each method with LLaMA-2-70B on 2 NVIDIA A100 GPUs. (T.Block: Transformer Block)
}
\label{table:wiki_ppl}
 \centering
  \footnotesize
\begin{tabular}{l | M{1.2cm} |  M{1.3cm} | c | c |  cccc | ccc}
\toprule
\rowcolor{Gray}
 ~ & \bf Pruning & ~ & ~ & \bf Throughput & \multicolumn{4}{c|}{\bf OPT} & \multicolumn{3}{c}{\bf LLaMA-2} \\
\rowcolor{Gray}
 \bf Method & \bf Unit & \bf Sparsity & \bf Tokens/s  & \bf Improve. &  \bf  6.7B & \bf  13B & \bf  30B & \bf  66B & \bf  7B & \bf  13B & \bf  70B \\
\midrule
\midrule
Dense		& - & 0\%  & 299 & 1.00$\times$ & 10.86	& 10.13	& 9.56	& 9.34		& 5.47	& 4.88	& 3.32 \\
\midrule
SparseGPT	& 2:4 & 50\%  & 293 & 0.98$\times$ &  14.16	& 12.98	& 10.92	& 10.09		& 10.79	& 8.75	& 5.70 \\
Wanda		& 2:4 & 50\%  & 293 & 0.98$\times$ &  15.93	& 15.57	& 13.38	& 12122.15	& 12.09	& 8.99	& 5.48 \\
DSnoT		& 2:4 & 50\%  & 293 & 0.98$\times$ &   16.21	& 15.16	& 12.39	& 11037.35	& 11.97	& 8.87	& 5.49 \\
\midrule
LLM-Pruner	& Channel & 20\% & 314 & 1.05$\times$ &  -		& -		& -		& -			& 10.58	& 8.56	& -  \\
SliceGPT	& Channel & 20\% & 314 & 1.05$\times$ &  11.59	& 10.73	& 9.93	& 9.60		& 6.87	& 6.01	& 4.44 \\
SliceGPT	& Channel & 25\% & 331 & 1.11$\times$ &  12.10	& 11.04	& 10.13	& 9.75		& 7.55	& 6.63	& 4.89 \\
SliceGPT	& Channel & 30\% & 343 & 1.15$\times$ &  12.73	& 11.49	& 10.39	& 9.94		& 8.59	& 7.44	& 5.44 \\
\midrule
\rowcolor{LightCyan}
SLEB		& T. Block & 10\% & 336 & 1.12$\times$ &  11.22	& 10.16	& 9.57	& 9.33	& 6.95	& 5.63	& 3.98 \\
\rowcolor{LightCyan}
SLEB		& T. Block & 20\% & 381 & 1.27$\times$ &  12.94	& 11.40	& 10.73	& 10.18		& 9.14	& 6.80	& 4.88 \\

\bottomrule
\end{tabular}
\end{table}

\newpage
\subsection{Dependency on Calibration Dataset}
\label{appendix:more_dependency}

We examine the influence of the calibration dataset on the perplexity results of pruned LLaMA-2 models.
We use calibration datasets consist of 128 sequences randomly sampled from WikiText-2 and C4 training dataset, respectively.
We compare perplexity results of SparseGPT, Wanda with 2:4 pruning, SliceGPT with 25\% sparsity, and SLEB with 20\% sparsity.

SliceGPT relies on principal component analysis of input activations to identify which neurons to prune for channel-wise pruning. 
This calibration method heavily relies on input values, without adequately considering information from the weight matrices, making its performance more dependent on the input data pattern. 
In contrast, SLEB, SparseGPT, and Wanda utilize information from pre-trained weight matrices of LLMs, showing less reliance on specific datasets. 
These methods utilize computation results from LLMs, which involve weight matrices at the layer (SparseGPT, Wanda) or network level (SLEB), to determine which components should be pruned. 
Thanks to utilization of weight matrix information, SLEB exhibits more stable perplexity results compared to SliceGPT.

\begin{table}[h]
\centering
\footnotesize
\caption{Perplexity results of different pruning methods depending on the calibration dataset}
\label{tab:dependency_full}
\begin{tabular}{c|c|c|c|c|c|c}
\toprule
\rowcolor{Gray} 
Model & Evaluation Set & Calibration Set & SLEB & SparseGPT & Wanda & SliceGPT \\ \midrule
\multirow{2}{*}{LLaMA-2-7B} & \multirow{2}{*}{WikiText-2} & WikiText-2 & 9.14 & 8.67 & 11.35 & 7.55 \\
 &  & C4 & 8.97 & 10.79 & 12.09 & 17.31 \\ \midrule
\multirow{2}{*}{LLaMA-2-7B} & \multirow{2}{*}{C4} & WikiText-2 & 12.32 & 14.73 & 16.22 & 32.74 \\
 &  & C4 & 11.26 & 13.54 & 15.57 & 12.43 \\ \midrule
\multirow{2}{*}{LLaMA-2-13B} & \multirow{2}{*}{WikiText-2} & WikiText-2 & 6.80 & 7.07 & 8.32 & 6.63 \\
 &  & C4 & 6.94 & 8.75 & 8.99 & 11.00 \\ \midrule
\multirow{2}{*}{LLaMA-2-13B} & \multirow{2}{*}{C4} & WikiText-2 & 9.42 & 11.92 & 12.62 & 29.86 \\
 &  & C4 & 9.40 & 11.39 & 12.47 & 10.75 \\ \midrule
\multirow{2}{*}{LLaMA-2-70B} & \multirow{2}{*}{WikiText-2} & WikiText-2 & 4.88 & 4.98 & 5.26 & 4.89 \\
 &  & C4 & 5.16 & 5.70 & 5.48 & 7.76 \\ \midrule
\multirow{2}{*}{LLaMA-2-70B} & \multirow{2}{*}{C4} & WikiText-2 & 7.31 & 8.42 & 8.23 & 20.03 \\
 &  & C4 & 7.26 & 8.16 & 8.10 & 8.11 \\ \bottomrule
\end{tabular}
\end{table}
\newpage
\subsection{Zero-shot tasks}
\label{appendix:more_zeroshot}

\begin{table}[h]
\caption{Accuracies (\%) on zero-shot tasks}
\label{table:zerishot_full_opt}
 \centering
  \fontsize{8.8}{8.8} \selectfont
\begin{tabular}{c | l | c | c | c | c | c | c | c}
\toprule
\rowcolor{Gray}
\bf Model     & \bf Method	& \bf Sparsity	& \bf PIQA	& \bf WinoGrande & \bf HellaSwag	& \bf ARC-e	& \bf ARC-c	& \bf Avg.  \\
\midrule
            & Dense		& -			& 76.39	& 65.19		 & 67.16	& 60.14	& 34.64	& 60.70 \\
           & SpareGPT	& 2:4 (50\%)		& 74.21	& 60.77		 & 57.25	& 53.03	& 29.44	& 54.94 \\
		& Wanda		& 2:4 (50\%)		& 71.76	& 60.22		 & 54.20	& 51.18	& 28.33	& 53.14 \\
OPT-6.7B 		& SliceGPT	& 25\%		& 70.35	& 60.62		 & 58.15	& 52.78	& 29.52	& 54.28 \\
 		& SliceGPT	& 30\%		& 68.61	& 60.69		 & 54.56	& 52.15	& 29.01	& 53.00 \\
\rowcolor{LightCyan}
\cellcolor{white!} 		& SLEB		& 10\%		& 76.61	& 64.72		 & 66.36	& 58.67	& 33.62	& 60.00 \\
\rowcolor{LightCyan}
\cellcolor{white!}	& SLEB		& 20\%		& 74.92	& 61.33		 & 62.13	& 57.07	& 32.59	& 57.61 \\
\midrule
            & Dense		& -			& 76.82	& 64.80		 & 69.81	& 61.87	& 35.67	& 61.79 \\
 		& SpareGPT	& 2:4 (50\%)		& 74.16	& 62.43		 & 59.18	& 56.27	& 31.74	& 56.76 \\
 		& Wanda		& 2:4 (50\%)		& 72.25	& 61.72		 & 57.97	& 53.96	& 29.69	& 55.12 \\
OPT-13B 		& SliceGPT	& 25\%		& 73.67	& 64.25		 & 63.28	& 60.52	& 34.64	& 59.27 \\
 		& SliceGPT	& 30\%		& 71.82	& 62.90		 & 60.66	& 58.80	& 32.94	& 57.42 \\
\rowcolor{LightCyan}
\cellcolor{white!}	& SLEB		& 10\%		& 76.12	& 65.51		 & 69.81	& 61.78	& 37.12	& 62.07 \\
\rowcolor{LightCyan}
\cellcolor{white!}	& SLEB		& 20\%		& 75.90	& 63.46		 & 66.87	& 59.60	& 34.56	& 60.08 \\
\midrule
            & Dense		& -			& 78.07	& 68.19		 & 72.27	& 65.24	& 38.23	& 64.40 \\
 		& SpareGPT	& 2:4 (50\%)		& 75.24	& 65.67		 & 65.10	& 59.76	& 34.04	& 59.96 \\
 		& Wanda		& 2:4 (50\%)		& 75.46	& 63.54		 & 63.41	& 60.14	& 31.91	& 58.89 \\
OPT-30B 		& SliceGPT	& 25\%		& 75.30	& 66.61		 & 69.42	& 63.55	& 35.67	& 62.11 \\
 		& SliceGPT	& 30\%		& 74.97	& 65.04		 & 68.15	& 63.55	& 34.64	& 61.27 \\
\rowcolor{LightCyan}
\cellcolor{white!}	& SLEB		& 10\%		& 77.64	& 68.75		 & 72.32	& 65.12	& 38.57	& 64.48 \\
   \rowcolor{LightCyan}
\cellcolor{white!}	& SLEB		& 20\%		& 76.93	& 67.40		 & 70.62	& 61.99	& 37.37	& 62.86 \\
\midrule
            & Dense		& -			& 79.82	& 68.90		 & 74.85	& 67.21	& 40.02	& 66.16 \\
 		& SpareGPT	& 2:4 (50\%)		& 77.75	& 66.22		 & 68.59	& 63.34	& 35.75	& 62.33 \\
 		& Wanda		& 2:4 (50\%)		& 50.65	& 50.51		 & 25.97	& 25.29	& 27.22	& 35.93 \\
OPT-66B		& SliceGPT	& 25\%		& 78.40	& 67.09		 & 73.33	& 67.89	& 39.16	& 65.17 \\
 		& SliceGPT	& 30\%		& 77.42	& 66.30		 & 72.62	& 66.90	& 37.97	& 64.24 \\
\rowcolor{LightCyan}
\cellcolor{white!}	& SLEB		& 10\%		& 78.94	& 68.90		 & 74.18	& 65.95	& 38.91	& 65.38 \\
\rowcolor{LightCyan}
\cellcolor{white!}	& SLEB		& 20\%		& 76.99	& 66.22		 & 70.77	& 63.34	& 35.32	& 62.53 \\
\bottomrule

\midrule
& Dense		& -			& 79.11	& 69.06			& 75.99		& 74.58	& 46.25	& 69.00	\\
& SpareGPT	& 2:4 (50\%)		& 72.14	& 64.96			& 58.93		& 60.90	& 34.22	& 58.23	\\
& Wanda		& 2:4 (50\%)		& 70.84	& 62.27			& 55.33		& 57.58	& 31.91	& 55.59	\\
LLaMA-2-7B & SliceGPT	& 25\%		& 66.87	& 63.38			& 54.16		& 58.46	& 34.56	& 55.49	\\
& SliceGPT	& 30\%		& 63.55	& 61.33			& 49.62		& 51.77	& 31.23	& 51.50	\\
\rowcolor{LightCyan}
\cellcolor{white!} & SLEB		& 10\%		& 76.44	& 63.14			& 70.23		& 63.68	& 37.71	& 62.24	\\
\rowcolor{LightCyan}
\cellcolor{white!} & SLEB		& 20\%		& 73.07	& 58.96			& 62.47		& 56.48	& 33.02	& 56.80	\\
\midrule
& Dense		& -			& 80.47	& 72.22			& 79.39		& 77.48	& 49.23	& 71.76	\\
& SpareGPT	& 2:4 (50\%)		& 75.46	& 68.51			& 65.52		& 66.04	& 39.76	& 63.06	\\
& Wanda		& 2:4 (50\%)		& 73.94	& 67.01			& 63.09		& 64.31	& 37.80	& 61.23	\\
LLaMA-2-13B & SliceGPT	& 25\%		& 68.55	& 67.48			& 58.10		& 62.50	& 37.88	& 58.90	\\
& SliceGPT	& 30\%		& 66.10	& 65.11			& 52.69		& 56.82	& 35.07	& 55.16	\\
\rowcolor{LightCyan}
\cellcolor{white!} & SLEB		& 10\%		& 79.16	& 66.93			& 74.36		& 71.84	& 41.55	& 66.77	\\
\rowcolor{LightCyan}
\cellcolor{white!} & SLEB		& 20\%		& 76.61	& 64.96			& 70.55		& 64.35	& 38.31	& 62.96	\\
\midrule
& Dense		& -			& 82.70	& 77.98			& 83.84		& 80.98	& 57.34	& 76.57	\\
& SpareGPT	& 2:4 (50\%)		& 80.03	& 76.56			& 76.09		& 76.94	& 49.74	& 71.87	\\
& Wanda		& 2:4 (50\%)		& 80.30	& 74.66			& 79.22		& 76.35	& 51.19	& 72.34	\\
LLaMA-2-70B & SliceGPT	& 25\%		& 74.92	& 75.37			& 68.84		& 77.90	& 51.71	& 69.75	\\
& SliceGPT	& 30\%		& 72.31	& 73.56			& 63.69		& 73.40	& 47.61	& 66.11	\\
\rowcolor{LightCyan}
\cellcolor{white!} & SLEB		& 10\%		& 81.56	& 75.06			& 80.02		& 76.77	& 52.30	& 73.14	\\
\rowcolor{LightCyan}
\cellcolor{white!} & SLEB		& 20\%		& 80.14	& 72.93			& 77.21		& 75.38	& 48.38	& 70.81	\\
\bottomrule

\end{tabular}
\vspace{-8mm}
\end{table}

\newpage
\subsection{Speedup}
\label{appendix:more_speedup}

We provide a comprehensive analysis of LLM inference speed across a range of LLM models using both NVIDIA A100 GPUs and A6000 GPUs.
For prompt processing, we conduct a single dummy test and then measure the latencies of 50 additional tests. The average latency is calculated from these measurements.
Similarly, for token generation, we conduct a single dummy test and then measure the throughputs of 10 additional tests. The average throughput is calculated from these measurements.

As presented in Table~\ref{tab:speedup_prompt_processing_full} and Table~\ref{tab:speedup_token_generation_full}, SLEB consistently enhances the latency and throughput of LLM inference across various serving scenarios.
However, the impact of other pruning methods that prune weight values is significantly influenced by the specific serving scenarios.
As discussed in Section~\ref{sec:pruning}, there are two reasons behind this.
First, the impact of weight pruning on the speed of matrix multiplication varies depending on the original matrix size, even when the pruning is conducted in a structured manner.
Second, LLMs have various types of operations besides linear operations with matrix multiplication, and these operations can also affect the overall inference speed.

\begin{table}[h]
\caption{LLaMA-2 latency for prompt processing}
\label{tab:speedup_prompt_processing_full}
\footnotesize
\centering
 \begin{tabular}{c | l M{1.0cm} | M{0.6cm}M{0.9cm}M{1.1cm} | M{0.6cm}M{0.9cm}M{1.1cm} | M{0.6cm}M{0.9cm}M{1.1cm} }
 \toprule
 \rowcolor{Gray}
  ~ & ~ & ~ & \multicolumn{3}{c|}{\bf 7B} & \multicolumn{3}{c|}{\bf 13B} & \multicolumn{3}{c}{\bf 70B} \\
 \rowcolor{Gray}
 & \bf Method & \bf Sparsity & \bf \#GPU & \bf Latency & \bf Speedup &\bf \#GPU & \bf Latency & \bf Speedup & \bf \#GPU & \bf Latency & \bf Speedup\\
\midrule
\midrule
&Dense & - & 1 & 240.0 & - & 1 & 397.3 &  - & 2 & 1718.4 & -\\
&2:4 Pruning & 50\% & 1 & 218.2 & 1.10$\times$ & 1 & 372.2 & 1.07$\times$ & 2 & 1555.5 & 1.10$\times$ \\
A100&Ch. Pruning & 25\% & 1 & 213.3 & 1.13$\times$ & 1 & 349.9 & 1.14$\times$ & 2 & 1440.7 & 1.19$\times$ \\
\rowcolor{LightCyan}
\cellcolor{white!}&SLEB & 10\% & 1 & 209.3 & 1.15$\times$ & 1 & 355.1 & 1.12$\times$ & 2 & 1529.1 & 1.12$\times$ \\
\rowcolor{LightCyan}
\cellcolor{white!}&SLEB & 20\% & 1 & 187.3 & 1.28$\times$ & 1 & 316.0 & 1.26$\times$ & 2 & 1364.1 & 1.26$\times$ \\
\bottomrule
\midrule
&Dense & - & 1 & 234.5 & - & 1 & 400.8 &  - & 4 & 1806.8 & -\\
&2:4 Pruning & 50\% & 1 & 216.7 & 1.08$\times$ & 1 & 371.7 & 1.08$\times$ & 4 & 1707.2 & 1.06$\times$ \\
A6000&Ch. Pruning & 25\% & 1 & 201.3 & 1.16$\times$ & 1 & 339.7 & 1.18$\times$ & 4 & 1475.6 & 1.22$\times$ \\
\rowcolor{LightCyan}
\cellcolor{white!}&SLEB & 10\% & 1 & 205.8 & 1.14$\times$ & 1 & 360.2 & 1.11$\times$ & 4 & 1616.2 & 1.12$\times$ \\
\rowcolor{LightCyan}
\cellcolor{white!}&SLEB & 20\% & 1 & 184.4 & 1.27$\times$ & 1 & 321.0 & 1.25$\times$ & 4 & 1434.3 & 1.26$\times$ \\
\bottomrule
\end{tabular}
\end{table}

\begin{table}[h]
\caption{LLaMA-2 throughput for token generation}
\label{tab:speedup_token_generation_full}
\footnotesize
\centering
 \begin{tabular}{c | l M{1.0cm} | M{0.6cm} M{1.0cm} M{1.1cm} | M{0.6cm}M{1.0cm}M{1.1cm} | M{0.6cm}M{1.0cm}M{1.1cm} }
 \toprule
 \rowcolor{Gray}
  ~ & ~ & ~ & \multicolumn{3}{c|}{\bf 7B} & \multicolumn{3}{c|}{\bf 13B} & \multicolumn{3}{c}{\bf 70B} \\
 \rowcolor{Gray}
 & \bf Method & \bf Sparsity & \bf \#GPU & \bf Tokens/s & \bf Improve. &\bf \#GPU & \bf Tokens/s & \bf Improve. & \bf \#GPU & \bf Tokens/s & \bf Improve.\\
\midrule
\midrule
&Dense & - & 1 & 1649 & - & 1 & 1078 &  - & 2 & 299 & -\\
&2:4 Pruning & 50\% & 1 & 1306 & 0.79$\times$ & 1 & 952 & 0.88$\times$ & 2 & 293 & 0.98$\times$ \\
A100&Ch. Pruning & 25\% & 1 & 1685 & 1.02$\times$ & 1 & 1109 & 1.03$\times$ & 2 & 331 & 1.11$\times$ \\
\rowcolor{LightCyan}
\cellcolor{white!}&SLEB & 10\% & 1 & 1856 & 1.13$\times$ & 1 & 1189 & 1.10$\times$ & 2 & 336 & 1.12$\times$ \\
\rowcolor{LightCyan}
\cellcolor{white!}&SLEB & 20\% & 1 & 2060 & 1.25$\times$ & 1 & 1337 & 1.24$\times$ & 2 & 381 & 1.27$\times$ \\
\bottomrule
\midrule
&Dense & - & 1 & 1200 & - & 1 & 712 &  - & 4 & 104 & -\\
&2:4 Pruning & 50\% & 1 & 1247 & 1.04$\times$ & 1 & 795 & 1.12$\times$ & 4 & 129 & 1.24$\times$ \\
A6000&Ch. Pruning & 25\% & 1 & 1335 & 1.11$\times$ & 1 & 827 & 1.16$\times$ & 4 & 135 & 1.29$\times$ \\
\rowcolor{LightCyan}
\cellcolor{white!}&SLEB & 10\% & 1 & 1362 & 1.14$\times$ & 1 & 783 & 1.10$\times$ & 4 & 117 & 1.13$\times$ \\
\rowcolor{LightCyan}
\cellcolor{white!}&SLEB & 20\% & 1 & 1515 & 1.26$\times$ & 1 & 879 & 1.24$\times$ & 4 & 135 & 1.30$\times$ \\
\bottomrule
\end{tabular}
\end{table}



\newpage
\subsection{Compatibility with Post-Training Quantization}
\label{appendix:ptq}

We demonstrate the compatibility of SLEB with PTQ.
When further compressing LLMs that have been pruned with SLEB using AWQ, we observe a negligible impact on perplexity results for both C4 and WikiText-2 datasets.
Additionally, it is notable that the impact of quantization is less pronounced on larger models with more than 13 billion parameters.

\begin{figure}[h]
    \centering
    \begin{subfigure}{0.45\columnwidth}
      \centering
      \includegraphics[width=\columnwidth]{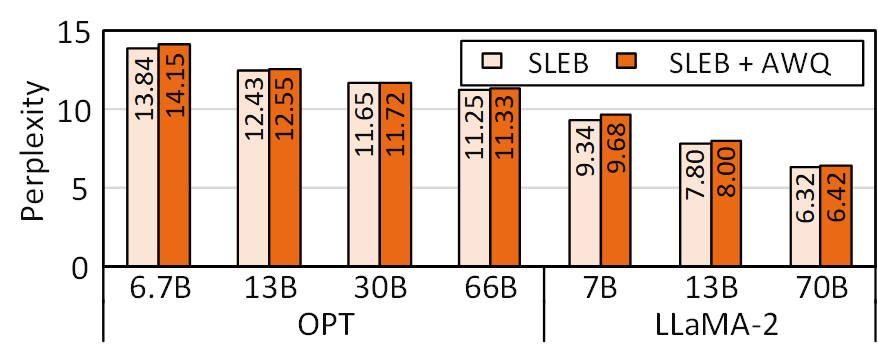}
      \vspace{-6mm}
      \caption{10\% SLEB \& 4-bit AWQ}
    \end{subfigure} \hspace{0.2cm}
    \begin{subfigure}{0.45\columnwidth}
      \centering
      \includegraphics[width=\columnwidth]{images/sleb_20_ptq_v2_c4.png}
      \vspace{-6mm}
      \caption{20\% SLEB \& 4-bit AWQ}
    \end{subfigure}
    \vspace{-3mm}
    \caption{Perplexity comparison between LLMs pruned with SLEB and those further compressed using AWQ, a 4-bit weight quantization (Evaluation dataset: C4)}
    \label{fig:sleb_ptq_c4_appendix}
\end{figure}

\begin{figure}[h]
    \centering
    \begin{subfigure}{0.45\columnwidth}
      \centering
      \includegraphics[width=\columnwidth]{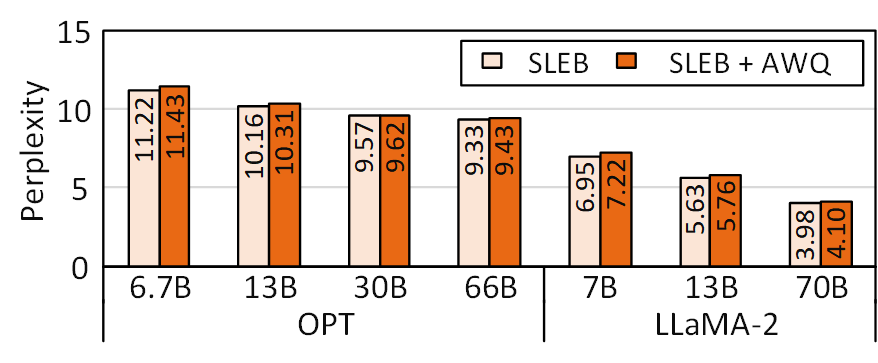}
      \vspace{-6mm}
      \caption{10\% SLEB \& 4-bit AWQ}
    \end{subfigure} \hspace{0.2cm}
    \begin{subfigure}{0.45\columnwidth}
      \centering
      \includegraphics[width=\columnwidth]{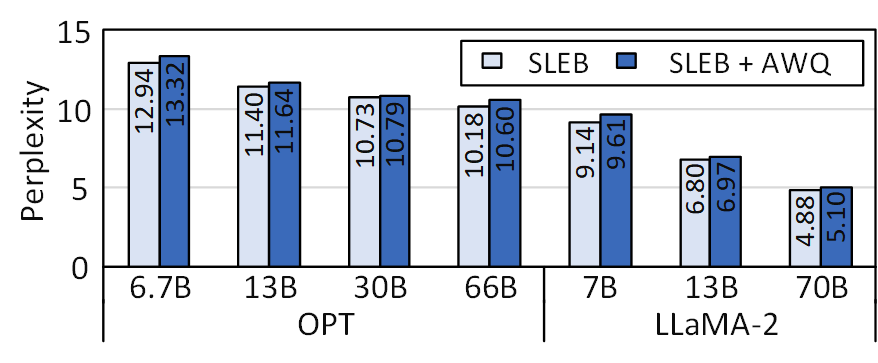}
      \vspace{-6mm}
      \caption{20\% SLEB \& 4-bit AWQ}
    \end{subfigure}
    \vspace{-3mm}
    \caption{Perplexity comparison between LLMs pruned with SLEB and those further compressed using AWQ, a 4-bit weight quantization (Evaluation dataset: Wikitext-2)}
    \label{fig:sleb_ptq_wiki_appendix}
\end{figure}

\newpage
\subsection{Perplexity and Accuracy Results of SLEB across Various Sparsity Ratios}
\label{appendix:ratios}

We explore SLEB with various sparsity ratios from 10\% to 50\%. We evaluate the perplexity with C4 dataset and the accuracy with zero-shot tasks. The evaluation results of OPT and LLaMA-2 models are provided in Table~\ref{tab:ratios_opt}.

\begin{table}[h]
\centering
\footnotesize
\vspace{-3mm}
\caption{OPT and LLaMA-2 perplexity (PPL) results on C4 dataset and accuracies on zero-shot tasks}
\label{tab:ratios_opt}
\begin{tabular}{c|c|c|c|c|c|c|c|c}
\toprule
\rowcolor{Gray}
 & & \bf PPL & \multicolumn{6}{c}{\bf Accuracy (\%)} \\
 \rowcolor{Gray}
\textbf{Model} & \textbf{Sparsity} & \textbf{C4 (↓)} & \textbf{PIQA} & \textbf{WinoGrande} & \textbf{Hellwswag} & \textbf{ARC-e} & \textbf{ARC-c} & \textbf{Avg.} \\ \midrule
 & Dense & 12.71 & 76.39 & 65.19 & 67.16 & 60.14 & 34.64 & 60.70 \\
 & 10\% & 13.84 & 76.61 & 64.72 & 66.36 & 58.67 & 33.62 & 60.00 \\
 & 20\% & 15.99 & 74.92 & 61.33 & 62.13 & 57.07 & 32.59 & 57.61 \\
 & 30\% & 21.11 & 71.76 & 58.25 & 54.28 & 52.27 & 28.07 & 52.93 \\
 & 40\% & 37.90 & 67.57 & 56.12 & 45.27 & 43.14 & 26.96 & 47.81 \\
\multirow{-6}{*}{OPT-6.7B} & 50\% & 135.49 & 61.48 & 51.30 & 32.73 & 35.86 & 23.38 & 40.95 \\ \midrule
 & Dense & 12.06 & 76.82 & 64.80 & 69.81 & 61.87 & 35.67 & 61.79 \\
 & 10\% & 12.43 & 76.12 & 65.51 & 69.81 & 61.78 & 37.12 & 62.07 \\
 & 20\% & 13.81 & 75.90 & 63.46 & 66.87 & 59.60 & 34.56 & 60.08 \\
 & 30\% & 17.13 & 73.07 & 58.48 & 62.29 & 54.97 & 33.11 & 56.38 \\
 & 40\% & 28.54 & 66.54 & 52.25 & 50.90 & 44.53 & 27.99 & 48.44 \\
\multirow{-6}{*}{OPT-13B} & 50\% & 106.66 & 58.87 & 51.38 & 34.59 & 34.64 & 25.60 & 41.02 \\ \midrule
 & Dense & 11.44 & 78.07 & 68.19 & 72.27 & 65.24 & 38.23 & 64.40 \\
 & 10\% & 11.65 & 77.64 & 68.75 & 72.32 & 65.12 & 38.57 & 64.48 \\
 & 20\% & 12.74 & 76.93 & 67.40 & 70.62 & 61.99 & 37.37 & 62.86 \\
 & 30\% & 15.59 & 74.81 & 62.98 & 65.33 & 59.51 & 32.76 & 59.08 \\
 & 40\% & 24.34 & 71.60 & 58.48 & 55.57 & 48.40 & 28.67 & 52.54 \\
\multirow{-6}{*}{OPT-30B} & 50\% & 47.27 & 63.93 & 53.67 & 42.44 & 38.68 & 25.00 & 44.74 \\ \midrule
 & Dense & 10.99 & 79.82 & 68.90 & 74.85 & 67.21 & 40.02 & 66.16 \\
 & 10\% & 11.25 & 78.94 & 68.90 & 74.18 & 65.95 & 38.91 & 65.38 \\
 & 20\% & 12.54 & 76.99 & 66.22 & 70.77 & 63.34 & 35.32 & 62.53 \\
 & 30\% & 15.81 & 74.97 & 60.22 & 64.46 & 57.03 & 33.87 & 58.11 \\
 & 40\% & 24.24 & 70.46 & 53.59 & 53.18 & 49.87 & 29.44 & 51.31 \\
\multirow{-6}{*}{OPT-66B} & 50\% & 58.09 & 60.50 & 48.14 & 36.32 & 36.15 & 24.40 & 41.10 \\ 
\bottomrule
\midrule
 & Dense & 7.26 & 79.11 & 69.06 & 75.99 & 74.58 & 46.25 & 69.00 \\
 & 10\% & 9.34 & 76.44 & 63.14 & 70.23 & 63.68 & 37.71 & 62.24 \\
 & 20\% & 12.32 & 73.07 & 58.96 & 62.47 & 56.48 & 33.02 & 56.80 \\
 & 30\% & 17.42 & 68.44 & 54.30 & 54.08 & 51.85 & 30.80 & 51.89 \\
 & 40\% & 32.06 & 62.73 & 52.17 & 42.75 & 42.89 & 28.75 & 45.86 \\
\multirow{-6}{*}{LLaMA-2-7B} & 50\% & 85.96 & 57.45 & 50.59 & 33.56 & 34.85 & 27.05 & 40.70 \\ \midrule
 & Dense & 6.73 & 80.47 & 72.22 & 79.39 & 77.48 & 49.23 & 71.76 \\
 & 10\% & 7.80 & 79.16 & 66.93 & 74.36 & 71.84 & 41.55 & 66.77 \\
 & 20\% & 9.42 & 76.61 & 64.96 & 70.55 & 64.35 & 38.31 & 62.96 \\
 & 30\% & 11.61 & 74.92 & 62.90 & 63.34 & 58.09 & 35.07 & 58.86 \\
 & 40\% & 16.35 & 70.95 & 59.91 & 55.54 & 51.64 & 30.72 & 53.75 \\
\multirow{-6}{*}{LLaMA-2-13B} & 50\% & 36.77 & 63.06 & 55.17 & 43.50 & 41.25 & 26.02 & 45.80 \\ \midrule
 & Dense & 5.71 & 82.70 & 77.98 & 83.84 & 80.98 & 57.34 & 76.57 \\
 & 10\% & 6.32 & 81.56 & 75.06 & 80.02 & 76.77 & 52.30 & 73.14 \\
 & 20\% & 7.31 & 80.14 & 72.93 & 77.21 & 75.38 & 48.38 & 70.81 \\
 & 30\% & 8.64 & 77.80 & 69.38 & 73.57 & 72.01 & 44.11 & 67.37 \\
 & 40\% & 12.16 & 76.01 & 65.90 & 65.48 & 66.84 & 37.89 & 62.42 \\
\multirow{-6}{*}{LLaMA-2-70B} & 50\% & 19.68 & 70.29 & 61.09 & 57.65 & 57.83 & 35.58 & 56.49 \\ 
\bottomrule
\end{tabular}
\end{table}

\clearpage
\subsection{Fine-tuning}
\label {fine-tuning}

We investigate the effects of fine-tuning on LLaMA-2-7B after pruning with 2:4 pruning (SparseGPT), channel-wise pruning (SliceGPT), and SLEB.
All pruned models are trained fore a single epoch on the Alpaca training dataset~\cite{alpaca}.
For the efficient fine-tuning procedure, we employ LoRA~\cite{lora} modules with a rank of 8.

Even after retraining, SLEB continues to outperform previous pruning methods in terms of both perplexity and zero-shot accuracy.

\begin{table}[h]
\centering
\footnotesize
\caption{Perplexity (PPL) and accuracy results before/after fine-tuning pruned LLaMA-2-7B}
\label{tab:fine-tuning}
\begin{tabular}{c|c|c|c|c|c|c|c|c|c}
\toprule
\rowcolor{Gray}
& & & \bf PPL & \multicolumn{6}{c}{\bf Accuracy (\%)}\\
\rowcolor{Gray}
\textbf{Method} & \textbf{Sparsity} & \textbf{Fine-tuning} & \textbf{C4 (↓)} & \textbf{PIQA} & \textbf{WinoGrande} & \textbf{Hellwswag} & \textbf{ARC-c} & \textbf{ARC-e} & \textbf{Avg.} \\ \midrule
Dense & 0\% & - & 7.26 & 79.11 & 69.06 & 75.99 & 46.25 & 74.58 & 69.00 \\
SparseGPT & 2:4 & - & 13.54 & 72.14 & 64.96 & 58.93 & 60.9 & 34.22 & 58.23 \\
SliceGPT & 20\% & - & 26.06 & 69.42 & 65.11 & 59.04 & 37.54 & 59.76 & 58.17 \\
SliceGPT & 25\% & - & 32.74 & 66.87 & 63.38 & 54.16 & 34.56 & 58.46 & 55.49 \\
SliceGPT & 30\% & - & 41.69 & 63.55 & 61.33 & 49.62 & 31.23 & 51.77 & 51.50 \\
\rowcolor{LightCyan} 
SLEB & 10\% & - & 9.17 & 76.77 & 61.33 & 67.19 & 36.77 & 64.86 & 61.38 \\
\rowcolor{LightCyan}
SLEB & 20\% & - & 11.38 & 74.27 & 58.80 & 62.08 & 32.94 & 59.34 & 57.49 \\
\midrule
SparseGPT & 2:4 & \checkmark & 10.65 & 75.95 & 66.30 & 68.03 & 38.40 & 64.35 & 62.61 \\
SliceGPT & 20\% & \checkmark & 13.50 & 73.88 & 64.16 & 69.3 & 40.02 & 62.29 & 61.93 \\
SliceGPT & 25\% & \checkmark & 14.79 & 71.7 & 62.83 & 65.93 & 38.82 & 61.65 & 60.19 \\
SliceGPT & 30\% & \checkmark & 16.35 & 69.86 & 61.64 & 62.37 & 37.37 & 59.76 & 58.20 \\
\rowcolor{LightCyan}
SLEB & 10\% & \checkmark & 9.01 & 78.73 & 62.43 & 70.70 & 41.21 & 66.25 & 63.86 \\
\rowcolor{LightCyan}
SLEB & 20\% & \checkmark & 10.50 & 73.88 & 64.16 & 69.30 & 40.02 & 62.29 & 61.07 \\ \bottomrule
\end{tabular}
\end{table}


\end{document}